\documentclass[sigconf]{acmart}

\usepackage{booktabs} 
\usepackage{subcaption}
\usepackage{amsmath}
\usepackage{array}
\usepackage{algorithm}
\usepackage[noend]{algpseudocode}
\usepackage{tabularx}
\usepackage{balance}

\newcolumntype{L}{>{\arraybackslash}m{2.8cm}}
\newcolumntype{R}{>{\arraybackslash}m{2.8cm}}

\setcopyright{rightsretained}

\acmDOI{10.475/123_4}

\acmISBN{123-4567-24-567/08/06}

\acmConference[KDD 2017]{Knowledge Discovery in Databases 2017}{August 2017}{Halifax, Nova Scotia, Canada} 
\acmYear{2017}
\copyrightyear{2017}

\begin{document}
\title{An Efficient Bandit Algorithm for Realtime Multivariate Optimization}

\author{Daniel N. Hill}
\affiliation{%
  \institution{Amazon.com, Inc.}
  \city{Palo Alto} 
  \state{CA} 
}
\email{daniehil@amazon.com}

\author{Houssam Nassif}
\affiliation{%
  \institution{Amazon.com, Inc.}
  \city{Seattle} 
  \state{WA} 
}
\email{houssamn@amazon.com}

\author{Yi Liu}
\affiliation{%
  \institution{Amazon.com, Inc.}
  \city{Seattle} 
  \state{WA} 
}
\email{yiam@amazon.com}

\author{Anand Iyer}
\affiliation{
  \institution{Amazon.com, Inc.}
  \city{Seattle} 
  \state{WA} 
}
\email{aiyer@amazon.com}

\author{S. V. N. Vishwanathan}
\affiliation{%
  \institution{Amazon.com, Inc. \& UC Santa Cruz}
  \city{Palo Alto} 
  \state{CA} 
}
\email{vishy@amazon.com}

\renewcommand{\shortauthors}{D. N. Hill et al.}

\begin{abstract}
Optimization is commonly employed to determine the content of web pages, such as to maximize conversions on
landing pages or click-through rates on search engine result pages. Often the layout of these pages can be decoupled into  
several separate decisions. For example, the composition of a landing page may involve deciding which image to show, 
which wording to use, what color background to display, etc. Such optimization is a combinatorial problem 
over an exponentially large decision space. Randomized experiments do not scale well to this setting, and 
therefore, in practice, one is typically limited to optimizing a single aspect of a web page at a time. This 
represents a missed opportunity in both the speed of experimentation and the exploitation of possible 
interactions between layout decisions.

Here we focus on multivariate optimization of interactive web pages. We formulate an approach where
the possible interactions between different components of the page are modeled explicitly.  We apply bandit methodology 
to explore the layout space efficiently and use hill-climbing to select optimal content in realtime. Our algorithm also
extends to contextualization and personalization of layout selection. Simulation results show the suitability
of our approach to large decision spaces with strong interactions between content. We further apply our algorithm 
to optimize a message that promotes adoption of an Amazon service. After only a single week of online optimization, 
we saw a 21\% conversion increase compared to the median layout. Our technique is currently being deployed 
to optimize content across several locations at Amazon.com.

\end{abstract}


\fancyhead{}

\copyrightyear{2017}
\acmYear{2017}
\setcopyright{rightsretained}
\acmConference{KDD '17}{August 13-17, 2017}{Halifax, NS, Canada}\acmDOI{10.1145/3097983.3098184} \acmISBN{978-1-4503-4887-4/17/08}

\maketitle

\section{Introduction And Background}

Web page design involves a multitude of distinct but interdependent
decisions to determine its layout and content, all of which are optimized for some business goal~\cite{nassif2016music, Teo2016airstream}.
For example, a search engine results page may be constructed from a set 
of query results, sponsored links, and query refinements, with the goal of maximizing
click-through rate. A landing page for an advertisement may consist of a sales
pitch containing separate components, such as an image, text blurb, and 
call-to-action button which are selected to promote conversions. Large-scale
data collection offers the promise of automatic optimization of these components.
However, optimization of large decision spaces also offers many challenges.

Separate optimization of each component of a web page may be sub-optimal. An image and a text
blurb that appear next to each other may interact or resonate in a way that cannot 
be controlled when selecting them independently. When multiple decisions are taken combinatorially, this leads to an exponential explosion in 
the number of possible choices. Even a small set of decisions can quickly lead
to 1,000s of unique page layouts. Controlled A/B tests are well suited to simple binary decisions, but 
they do not scale well to hundreds or thousands of treatments. Finding the best layout is further
complicated by the need for contextualization and personalization which compounds the number of
factors that need to be considered simultaneously. A final challenge is that any solution for optimizing page layout
 needs to be deployed in a system that can make selections from the layout space 
 in realtime where latencies of only 10s of milliseconds may be acceptable.
 
One approach for efficiently learning to optimize a large decision space is
 fractional factorial design~\cite{joseph2006experiments,box2005statistics}. Here, a randomized experiment is designed to test only a fraction of the decision space. 
 Assumptions are made about how choices interact in order to infer the results for the untested treatments.
  Experimentation is accelerated with the caveat that higher-order interactions are aliased onto
 lower-order effects. In practice, higher-order interactions are often negligible so that this approximation is appropriate.
 However, these experiments suffer from their rigidity. The experimental designs follow a schedule that make it difficult to test new 
 ideas ad hoc. The experiments are also typically non-adaptive so that losing treatments cannot be 
 automatically suppressed. Fractional factorial designs also make no account of context.
 
A major alternative in fast experimentation is multi-armed bandit methodology. This class of algorithms balances
 exploration and exploitation to minimize regret, which represents the opportunity cost incurred while testing sub-optimal decisions 
before finding the optimal one. Bandit algorithms are effective for rapid experimentation because they concentrate 
testing on actions that have the greatest potential reward.  Bandit methods have shown impressive empirical 
performance~\cite{chapelle2011empirical} and can easily incorporate context~\cite{li2010contextual,Agrawal2013ContextualTSBounds}. There is also literature on 
applying bandit methods to combinatorial problems. This includes combinatorial bandits~\cite{cesa2012combinatorial} for subset selection.  Submodular bandits 
also select subsets while considering interactions between retrieved results to maintain diversity~\cite{yue2011linear, Teo2016airstream}. Thus, bandit algorithms are good candidates to efficiently 
discover the optimal assignment of content to a web page.

Here we present an approach to layout optimization using bandit methodology. We name our approach multivariate
bandits to distinguish from combinatorial bandits which are used to optimize subset selection. We propose a parametric
Bayesian model that explicitly incorporates interactions between components of a page. We avoid a combinatorial
explosion in model complexity by only considering pairwise interactions between page components. Contextualization and 
personalization are enabled by further allowing for pairwise interactions between content and context. We
efficiently balance exploration and exploitation through the use of Thompson sampling~\cite{agrawal2012analysis}.
We allow for realtime search of the layout space by applying greedy hill climbing during selection. 

Our approach is most similar to~\cite{wang2016beyond} where the authors used a model of
 pairwise interactions to optimize whole-page presentation. However, their algorithm addresses a different use-case of assigning content to the page's components where every piece of content
 is eligible for every slot. An alternative (and slower) approach to hill-climbing in the discrete decision space is to search in an equivalent continuous decision space that can be obtained as the convex hull of the binary decision vectors. A gradient descent approach yields a global-optimal vector that can then be rounded to output a binary decision vector.  In the online bandits setting, this translates to the approach taken by the authors in~\cite{mohan2011bipartite} and~\cite{cesa2012combinatorial}. Our approach is a greedy alternating optimization strategy that can run online in real-time.

\begin{figure}[h]
  \centering
  \includegraphics[width=\linewidth]{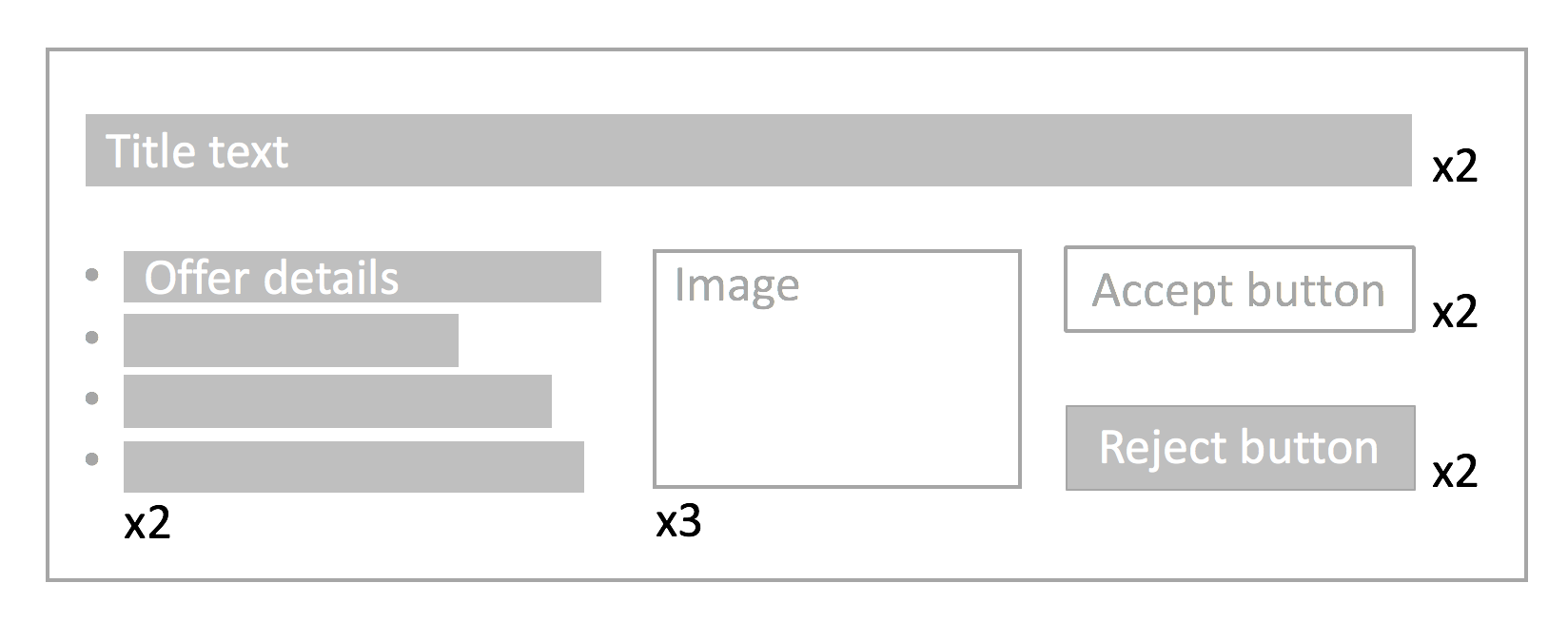}
  \caption{Example of a generic promotional message for an Amazon service. Each component is a separate widget with the indicated number of alternative content. There are 48 total distinct layouts.}
  \label{fig:desktop_example}
\end{figure}

Our solution has been deployed to a live production system to combinatorially optimize a landing page that promotes purchases of an Amazon service (Fig.~\ref{fig:desktop_example}),
leading to a 21\% increase in purchase rate. In the sections below, we describe our problem formulation, present our algorithm, demonstrate its 
properties on synthetic data, and finally analyze its performance in a live experiment.

\section{Problem Formulation}
\subsection{Problem setting}
We formally define the problem we address as the selection of a layout $A$ of a web page under a context $X$ in order to maximize the expected value of a reward $R$. The reward $R$ corresponds to the value of an action taken by a user after viewing the web page. $R$ could represent a click, signup, purchase or purchase amount. 

We assume that each possible layout $A$ is generated from a common template that contains $D$ widgets or slots representing the contents of the page. The $i^{th}$ widget has $N_i$ alternative variations for the content that can be placed there. Our approach is general and can handle a different number of variations per widget; however, for simplicity of formulation, we will assume that all widgets have the same number of possible variations, so that $N_1=N_2=\dots=N$.  Thus, a combinatorial web page has $N^D$ possible layouts. We represent a layout as $A \in \{1,2,\ldots,N\}^D$, a $D$-dimensional vector. $A[i]$ denotes the content chosen for the $i^{th}$ widget. 

Context $X$ represents user or session information that may impact a layout's expected reward. For example, context may include time of day, device type, or user history. We assume $X$ is drawn from some fixed unknown distribution. $X$ and $A$ are combined (possibly nonlinearly) to form the final feature vector $B_{A,X} \in \mathbb{R}^M$ of length $M$. $B_{A,X}$ can thus include interactions between an image displayed in a slot, the user's age, and time of day.

The reward $R_{A,X}$ for a given layout and context depends on a linear scaling of $B_{A,X}$ by a fixed but unknown vector of weights $\mu  \in \mathbb{R}^M$. We model the reward with a generalized linear model:
\begin{equation}\label{eq:GLM}
 \mathbb{E}[R|A,X] = g(B_{A,X}^\top \mu),
\end{equation}
where $g$ is the link function and $\top$ denotes the matrix transpose.

We now define a stochastic contextual multi-armed bandit problem~\cite{bubeck2012BanditsSurvey}. We have $N^D$ arms, one per layout. The algorithm proceeds in discrete time steps $t=1,2,\dots,T$. On trial $t$, a context $X_t$ is generated and a vector $B_{A,X_t}$ is revealed for each arm $A$. Note that whenever $t$ is a subscript, it indicates a time step index. An arm $A_t$ is selected by the algorithm and a reward $R_t=R_{A_t,X_t}$ is observed.

Let $\mathcal{H}_{t-1} = \{(A_\tau, R_\tau, X_\tau), \;\tau = 1,\dots, t-1\}$ represent the history prior to time $t$. Let $A^*_t$ denote the optimal arm at time $t$:
\begin{equation}
A^{*}_t = \underset{A}{argmax}\ E[R_{A,X_t}].
\label{eq:BestArmFromW}
\end{equation}
Let $\Delta_t$ be the difference between the expected reward of the optimal arm and the selected arm at time $t$:
\begin{equation}
	\Delta_t =   E[R_{A^*_t,X_t}]- E[R_{A_t,X_t}]. 
\end{equation}
The objective of our bandit problem is to estimate $\mu$ while minimizing cumulative regret over $T$ rounds:
\begin{equation}\label{eq:cumulativeRegret}
\Delta^T =  \sum_{t=1}^T \Delta_t. 
\end{equation}

The contextual bandit with linear reward problem has been well studied~\cite{bubeck2012BanditsSurvey}. \citeauthor{Dani2008ContextualBanditBound} established a theoretical lower bound of $\Omega(M\sqrt{T})$ for the regret, along with a matching upper bound~\cite{Dani2008ContextualBanditBound}. With an oracle which efficiently solves equation~\ref{eq:BestArmFromW}, then both~\cite{Dani2008ContextualBanditBound} and~\cite{Agrawal2013ContextualTSBounds} give an efficient algorithm with a regret bound of $\tilde{O}(M^{3/2}\,\sqrt{T})$, the best achieved by a computationally efficient algorithm. Efficient here means a polynomial number of calls to the oracle. In this work, we take care  in constructing an efficient oracle which seeks to approximate the solution to equation~\ref{eq:BestArmFromW} efficiently in our contextual setting.

Next we will specify how to compute and update a model of $\mu$, generate feature vector $B_{A,X}$ from layout $A$ and context $X$, and approximate equation~\ref{eq:BestArmFromW} in a combinatorial setting.

\subsection{Probability model}
Here we consider the case where reward $R$ is binary, although our approach can be extended to categorical and numeric outcomes. Let $R = +1$ indicate the user took the desired action and $R = -1$ indicate otherwise. We choose the probit function as our link function for equation~\ref{eq:GLM}. Our probit regression has binary reward probability:
\begin{equation} 
P(R | A, X) = \Phi \left(R * B_{A,X}^\top W\right),
\label{eq:BLIP}
\end{equation}
where $W$ is our estimate of $\mu$, and $\Phi$ is the cumulative distribution function of the standard normal distribution. We model the regression parameters $W$ as mutually independent random variables following a Gaussian posterior distributions, with Bayesian updates over an initial Gaussian prior of $\mathcal{N}(0,1)$. We update $W$ using observations $R$ and $B_{A,X}$ as described in~\cite{graepel2010web}. Note that in the remainder of the paper we will assume binary features so that $B_{A,X} \in [0,1]^M$. This is done for notational convenience, but it is straight-forward to extend our formulation to continuous inputs.

To capture all possible interactions between web page widgets, the number of model parameters would be $O(N^D)$. We avoid this combinatorial explosion by capturing only pair-wise interactions, and assuming that higher-order terms contribute negligibly to the probability of success. Ignoring context $X$, we denote the non-contextual feature vector as $B_{A}$, and our linear model becomes:
\begin{equation}\label{eq:NonContextualModel} 
B_A^\top W  = W^0 + \sum_{i=1}^{D}W^1_i(A) + \sum_{j=1}^{D}\sum_{k=j+1}^{D}W^2_{j,k}(A)
\end{equation}
where $W^0$ is a common bias weight, $W^1_i(A)$ is a weight associated with the content in the $i^{th}$ widget, and $W^2_{j,k}(A)$ is a weight for the interaction between the contents of the $j^{th}$ and $k^{th}$ widgets. $W$ thus contains $O(N^2D^2)$ terms (see Table~\ref{table:weights}).

\begin{table}[t]
  \caption{Definition of weights in models.}
  \label{table:weights}
  \centering
    {\renewcommand{\arraystretch}{1.4}
    \begin{tabularx}{\columnwidth}{|| c | X ||} \hline
      \textbf{Weight class} &  \textbf{Definition} \\ \hline
     $W^0$  & Bias weight  \\ \hline
     $W^1_i(A) $  & Impact of content in $i^{th}$ widget \\ \hline
     $W^2_{i,j}(A) $  & Interaction of content in $i^{th}$ widget with content in $j^{th}$ widget  \\ \hline
     $W^c_i(X) $  & Impact of $i^{th}$ contextual feature  \\ \hline
     $W^{1c}_{i,j}(A,X) $  & Interaction of content in $i^{th}$ widget with the $j^{th}$ contextual feature \\ \hline
     $W^L(A) $  & Weight associated with distinct layout A  \\ \hline
    \end{tabularx}
    }
\end{table}

To account for contextual information $X$ and possible interactions between web page content and context, additional terms can be added to $B_{A,X}$. We represent $X$ as a multidimensional categorical variable of dimension $L$ where each dimension can take one of $G$ values. Let $X_l$ represent the $l^{th}$ feature of the context. We add first-order weights for $X$ as well as second-order interactions between $A$ and $X$ features as
\begin{equation}\label{eq:ContextualModel} 
\begin{split}
B_{A,X}^\top W =& W^0 + \sum_{i=1}^{D}W^1_i(A) + \sum_{j=1}^{D}\sum_{k=j+1}^{D}W^2_{j,k}(A)\\
&+ \sum_{l=1}^L W^c_l(X) + \sum_{m=1}^D\sum_{n=1}^L W^{1c}_{m,n}(A,X)
\end{split}
\end{equation}
where $W^c_l(X)$ is a weight associated with the $l^{th}$ contextual feature and $W^{1c}_{m,n}(A,X)$ is a weight for the interaction between the content of the $m^{th}$ widget and the $n^{th}$ contextual feature.
$W$ now contains $O(NDGL + N^2D^2)$ terms.

\section{Multvariate Testing Algorithm}
Although our feature representation limits model complexity to $2^{nd}$-order interactions, we need an efficient method to learn the still large number of parameters. Furthermore, computing the $argmax$ in equation~\ref{eq:BestArmFromW} requires a search through $N^D$ layouts. We now describe a bandit strategy that efficiently searches for near optimal solutions without evaluating the entire space of possible layouts.

\subsection{Thompson Sampling}
Thompson sampling~\cite{chapelle2011empirical} is a common bandit algorithm used to balance exploration and exploitation in decision making. In our setting, this implies the user is not always shown the layout with the currently highest expected reward, but is also shown layouts that have high uncertainty and thus a potentially higher reward. Thompson sampling selects a layout proportionally to the probability of that layout being optimal conditioned on previous observations:
\begin{equation} 
A_t \sim P(A=A^{*}|X, \mathcal{H}_{t-1}). 
\end{equation}

In practice, this probability is not sampled directly. Instead one samples model parameters from their posterior and picks the layout that maximizes the reward, as in algorithm~\ref{alg:TS}. Note that weights $W_t$ are estimated from history $\mathcal{H}_{t-1}$. In our Bayesian linear probit regression, the model weights are represented by independent Gaussian random variables~\cite{graepel2010web} and so can be sampled efficiently. 

Let $\tilde{W} ~ P(W|\mathcal{H})$ be the sampled weights. The sampled reward probability is monotonic with $B_{A,X}^\top \tilde{W}$,  which is itself a Gaussian-distributed random variable. This property ensures that Thompson Sampling remains computationally efficient as long as we can efficiently solve $argmax_A\ B_{A,X}^\top \tilde{W}$~\cite{Agrawal2013ContextualTSBounds}.

\begin{algorithm}
\caption{Thompson Sampling for Contextual Bandits}\label{alg:TS}
\begin{algorithmic}[1]
  \ForAll{$t = 1, \ldots, T$} 
  \State Receive context $X_t$
  \State Sample $\tilde{W_t}$ from the posterior $P(W|\mathcal{H}_{t-1})$
  \State Select $A_t = argmax_A\ B_{A,X_t}^\top \tilde{W_t}$ \label{line:TS:argmax}
  \State Display layout $A_t$ and observe reward $R_t$
  \State Update $\mathcal{H}_{t} = \mathcal{H}_{t-1} \cup (A_t, R_t, X_t)$
  \EndFor
\end{algorithmic}
\end{algorithm}

In line~\ref{line:TS:argmax} of algorithm~\ref{alg:TS}, finding the best layout $A$ given $\tilde{W}$ and $B_{A,X}$ is an instance of the maximum edge-weighted clique problem, which is NP-Hard~\cite{macambira2000edgeWeightedClique}. At each round, it requires evaluating $O(N^D)$ layouts. Next we describe an efficient approximation.

\subsection{Hill climbing optimization}
Instead of an exhaustive search to find the true $argmax$ over $A$, we approximate it by greedy hill climbing optimization\cite{casella2002statistical}. We start by randomly picking a layout $A^0$. On each round $k$, we randomly choose a widget $i$ to optimize, while fixing the content of all other widgets. We cycle through the $N$ possible alternatives for widget $i$, and select content $j^*$ that maximizes the layout score:
\begin{equation}
j^* = \underset{j}{argmax} B_{A^{k-1}\leftarrow (A[i]=j),X}^\top \tilde{W} 
\end{equation}
where $A^{k-1}\leftarrow (A[i]=j)$ denotes layout $A^{k-1}$ updated so that widget $i$ is assigned content $j$. We then use the optimal content $j^*$ to generate $A^{k}$ from $A^{k-1}$.  We repeat this procedure $K$ times, each iteration optimizing the content of a single widget conditioned on the rest of the layout. If each widget is visited without the update procedure changing the content, that means the search has reached a local optimum and can be terminated early.

We thus replace line~\ref{line:TS:argmax} of the Thompson sampling algorithm (algorithm~\ref{alg:TS}) by a call to the hill climbing algorithm (algorithm~\ref{alg:HillClimbing}). Hill climbing potentially converges to a local optimum while evaluating $KN$ layouts. We alleviate the local optimum problem by performing random restarts~\cite{casella2002statistical}. We perform hill climbing $S$ times where each iteration $s$ uses a different initial random layout $A^0_s$. We return the best layout among the $S$ hill climbs, resulting in a maximum of $SKN$ layout evaluations. 

\begin{algorithm}
\caption{Hill climbing with random restarts}\label{alg:HillClimbing}
\begin{algorithmic}[1]
  \Function{Hill Climbing Search}{$\tilde{W}$, $X$}
  \For{$s=1,\ldots, S$}
  \State Pick a layout $A^0_s$ randomly
  \For{$k=1,\ldots, K$}
  \State Randomly choose a widget $i$ to optimize \label{line:HC:NextWidget}
  \State Find $j^* = argmax_j\, B_{A^{k-1}_s\leftarrow (A[i]=j),X}^\top \tilde{W}$ \label{line:HC:BestContent}
  \State $A^k_s = A^{k-1}_s\leftarrow (A[i]=j^*)$
  \EndFor
  \EndFor
  \State $s^* = argmax_s\, B_{A^K_s}^\top\tilde{W}$ 
  \State \Return $A^K_{s^*}$ \label{line:HC:BestContent}
  \EndFunction
\end{algorithmic}
\end{algorithm}

When we sequentially compare two layouts that differ by only one piece of content, we only need to sum $O(D+L)$ weights as the other weights in equation~\ref{eq:ContextualModel} are unchanged. Hill climbing combined with Thompson sampling and our probability model yield a time complexity of $O(SKN(D+L))$, compared with $O(N^D(D+L))$ for an exhaustive search.

\section{Simulation Results}

We first test our algorithm on a simulated data set. Our goal is to understand the algorithm's performance as we vary 
parameters of the simulated data relating to the (a) strength of interaction between slots, (b) complexity of the template 
space, and (c) importance of context. We refer to the non-contextual version of our multivariate algorithm as MVT2 (see Table \ref{table:algorithms}) where we use the representation described in 
equation~\ref{eq:NonContextualModel}. We also test the contextual version of this algorithm MVT2c which uses the representation in equation~\ref{eq:ContextualModel}.

\begin{table*}[t]
  \caption{Symbols for algorithms discussed in this paper.}
  \label{table:algorithms}
  \centering
  {\renewcommand{\arraystretch}{1.3}
    \begin{tabular}{|| c | l | c | c ||} 
     \hline
      \textbf{Algorithm} & \textbf{Description} & \textbf{\# Parameters} & \textbf{Equation} \\ \hline
       MVT1 & Probit model without interactions between widgets & $O(ND)$ & (\ref{eq:mvt1_lm}) \\ \hline
       MVT2 & Probit model with interactions between widgets  &   $O( N^2 D^2) $ & (\ref{eq:NonContextualModel}) \\ \hline
       MVT2c & Probit model with interactions between widgets and between widgets and context & $O(NDGL + N^2 D^2)$ & (\ref{eq:ContextualModel}) \\ \hline 
       $N^D$-MAB & Non-contextual multi-armed bandit with $N^D$ arms & $O(N^D)$ & (\ref{eq:ts_lm}) \\ \hline
       D-MABs  & Independent non-contextual N-armed bandit for each of D widgets & $O(ND)$ & (\ref{eq:d_mabs}) \\ \hline
    \end{tabular}
    }
\end{table*}

We also compare MVT2 to two baseline models. The first model, $N^D$-MAB, is a non-contextual multi-armed bandit where a layout is represented only by 
an identifier so that no relationship between layouts can be learned. Such a model has been applied previously to ad layout optimization~\cite{tang2013automatic}. We use a Bayesian linear probit regression where the linear model is: 
\begin{equation} 
\label{eq:ts_lm}
B_A^TW = W^L(A),
\end{equation}
with $W^L(A)$ being a weight associated with a distinct layout $A$. This baseline gives us a vanilla implementation of a multi-armed bandit algorithm with $N^D$ arms. 

A second baseline model MVT1
is obtained by dropping the 2nd order terms from MVT2. The linear model becomes: 
\begin{equation} 
\label{eq:mvt1_lm}
B_A^TW = W^0 + \sum_{i=1}^{D}W^1_i(A).
\end{equation}
The MVT1 model helps us evaluate the benefit of modeling interactions between the content of different widgets.

\subsection{Simulated data}
We generate simulated data consistent with the assumptions of the MVT2c model. We sample outcomes from the linear probit function 
of equation~\ref{eq:BLIP}. For its linear model we use:
\begin{equation} 
\label{simulator_lm}
\begin{split}
B_{A,X}^TW = &\  \frac{1}{\beta} [W^0 + \alpha_1\sum_{i=1}^{D}W^1_i(A)\\ 
&+ \alpha_2\sum_{j=1}^{D}\sum_{k=j+1}^{D}W^2_{j,k}(A) \\
&+ \alpha_c\sum_{l=1}^L W^c_l(X) + \alpha_c\sum_{m=1}^D\sum_{n=1}^L W^{1c}_{m,n}(A,X)],
\end{split}
\end{equation}
where $\beta$ is a scaling parameter and $\alpha_1$, $\alpha_2$, and $\alpha_c$ are control parameters. We manually set the control parameters
to define the relative importance of content, interactions between content, and context, respectively. The parameter $\beta$ is set so that the
overall variance of $B_{A,X}^TW$ is constant, ensuring the signal-to-noise ratio is equal across experiments. 
 For each simulation, the $W$ parameters are independently sampled from a normal distribution with mean 0 and variance 1. We use a context $X$  that is 
 univariate and uniformly distributed on the set of integers $\{1,2,\dots,G\}$. 

\begin{figure}[h]
   \includegraphics[width=\columnwidth]{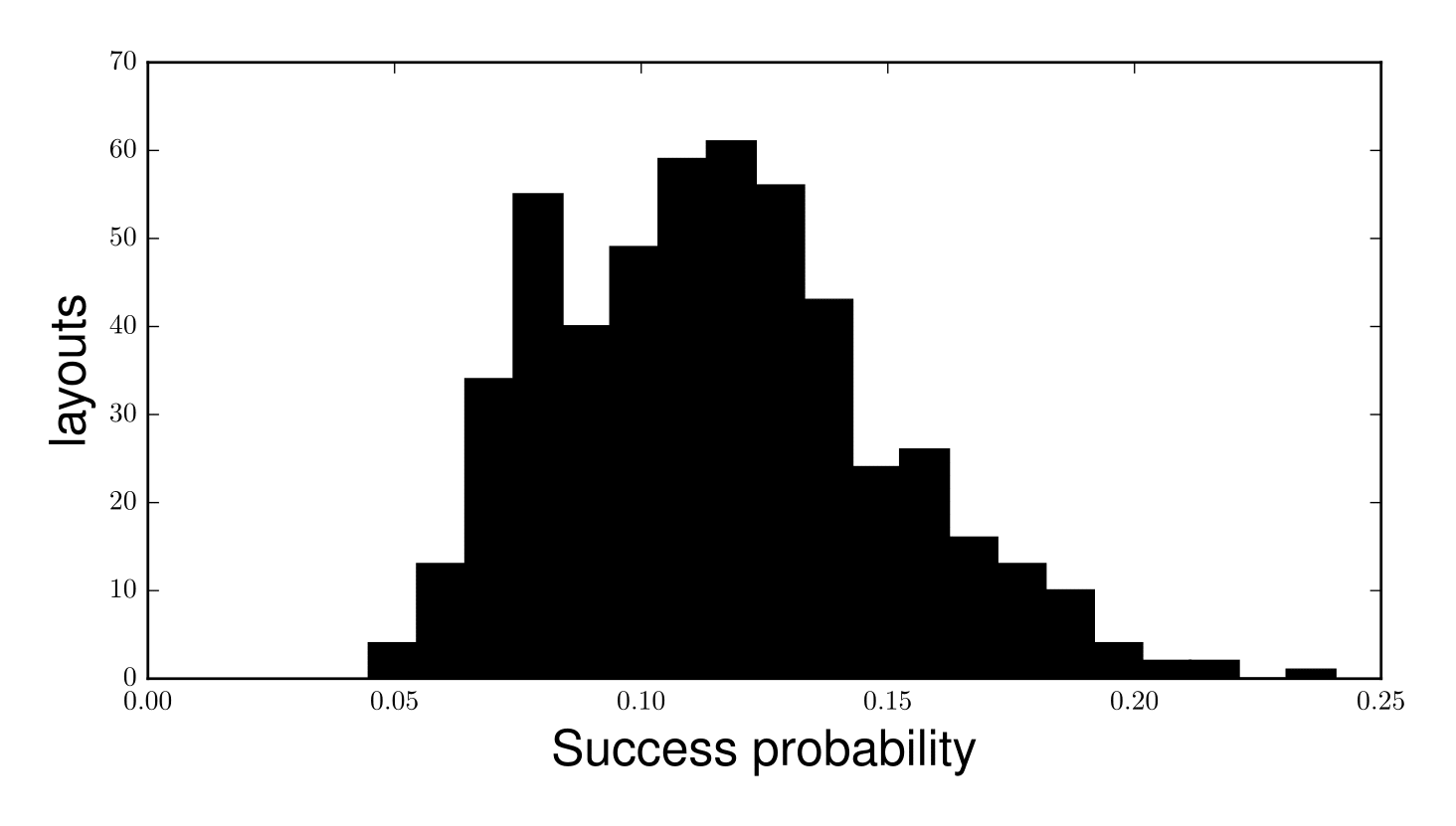}
   \label{fig:sim-ex-regret}
   \caption{Histogram of expected reward for typical set of simulated layouts.}
   \label{fig:simulation_histogram}
\end{figure}

Unless otherwise specified, we set $D=3$ and $N=8$, which yields $512$ possible layouts. Simulations were run for $T=250,000$ time steps. 
On each iteration of the simulation, a context is sampled at random and presented 
to the algorithm. A layout is chosen by applying algorithm~\ref{alg:TS}. Then, a binary outcome is sampled from the data generation model 
given the context and selected layout. The models are batch trained every 1000 iterations to simulate delayed feedback typically present in production systems. Each simulation
was run for 15 repetitions and plotted with standard errors. Figure~\ref{fig:simulation_histogram} shows the distribution of success probabilities for layouts in a typical simulation experiment.

We evaluate each model in terms of regret. We define empirical regret as the difference between the expected value 
of the optimal strategy and the empirical performance of the algorithm:
\begin{equation} 
\label{eq:regret}
regret = \frac{1}{T}\sum_{t=1}^{T}E(R|A_t^*,X_t) - R_t.
\end{equation}

We also define a $local \, regret$ by averaging the regret over a moving window with bounds $t_0$ and $t_1$ as:
\begin{equation} 
\label{eq:local_regret}
local \,regret[t_0,t_1] = \frac{1}{1+t_1-t_0}\sum_{t=t_0}^{t_1}E(R|A_t^*,X_t) - R_t.
\end{equation}

\subsection{Simulation experiments}
\begin{figure}[h]
   \centering
   \includegraphics[width=\columnwidth]{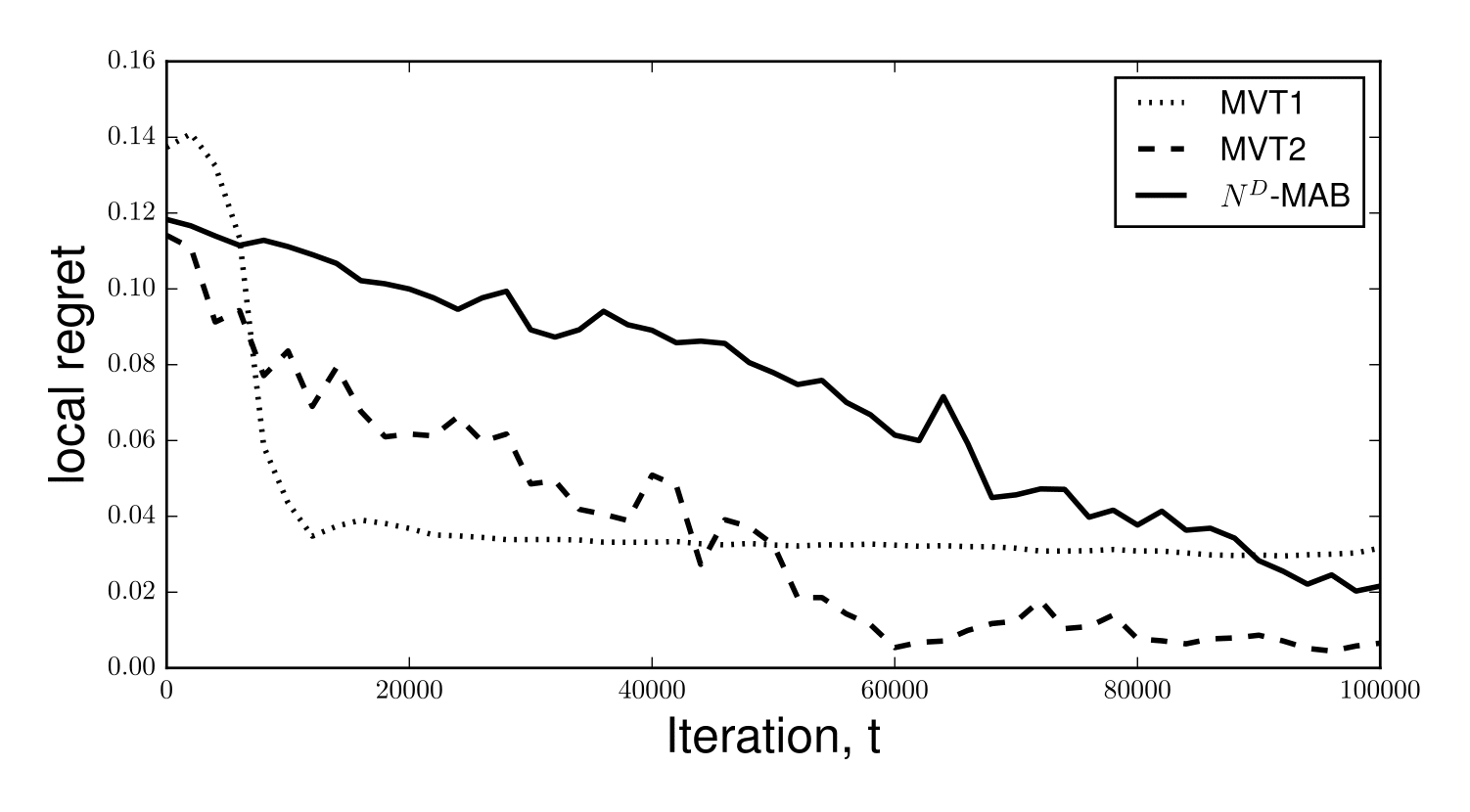}
   \caption{Example run of algorithms on simulated data with $\alpha_1=1$, $\alpha_2=2$, and $\alpha_c=0$. Local regret values are averaged over a moving window of 2500 iterations.}
   \label{fig:sim_example}
\end{figure}

First, we test the impact of varying the strength of interactions between layout widgets. We vary the amplitude of $\alpha_2$ while
fixing $\alpha_1=1$ and $\alpha_c=0$. An example simulation run for each algorithm is shown in figure~\ref{fig:sim_example}
with $\alpha_2=2$. We see that MVT1 converges quickly because of its smaller parameter set. However, its local regret plateaus 
to a higher value due to its inability to learn interactions between content. MVT2 and $N^D$-MAB are both able to nearly 
eliminate regret, though MVT2 converges faster due to better generalization between layouts.

\begin{figure}[h]
\centering
   \includegraphics[width=\columnwidth]{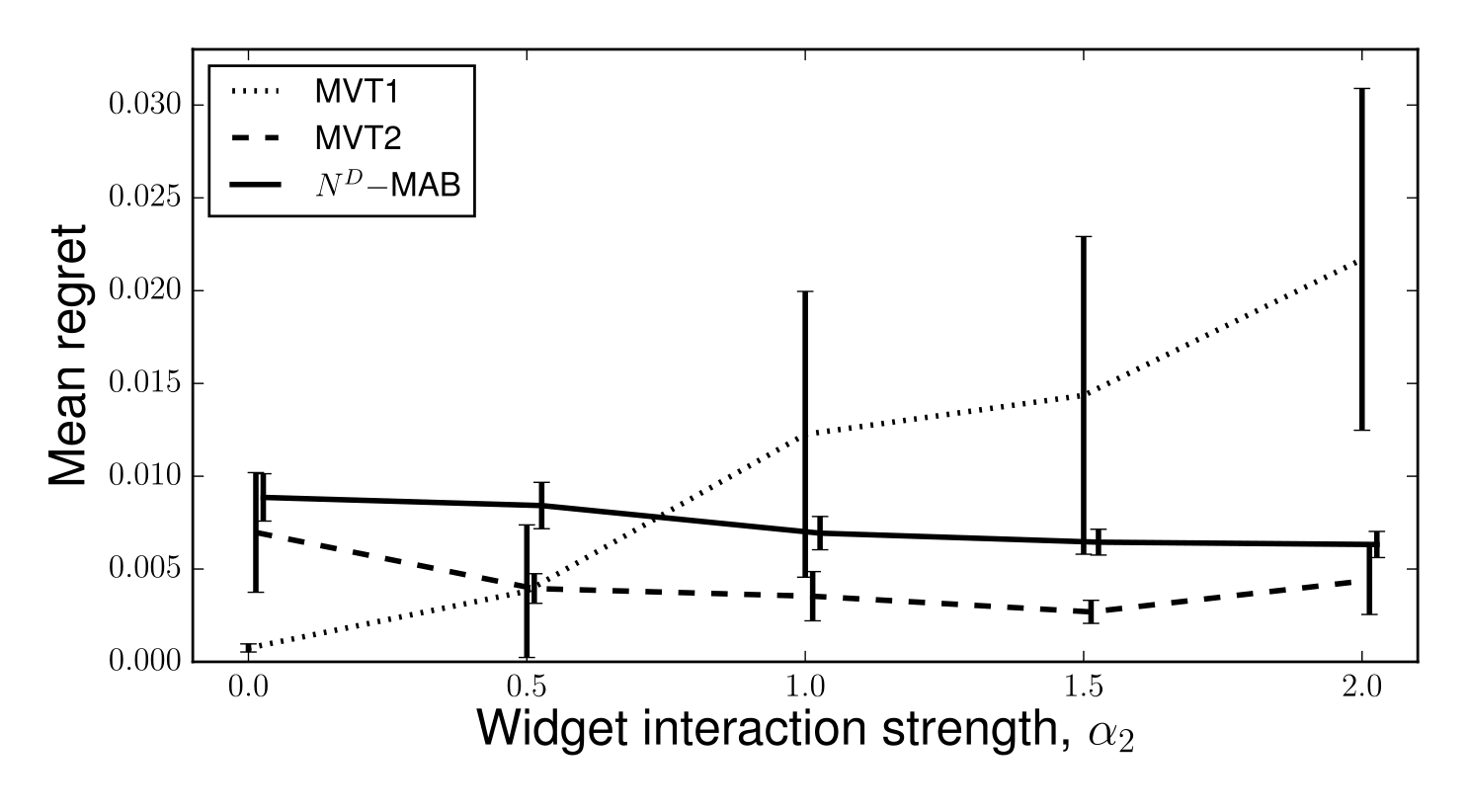}
\caption{Algorithm performance as $\alpha_2$ is varied.}
\label{fig:interaction_sim}
\end{figure}

As $\alpha_2$ is varied, we see this pattern continue (Fig.~\ref{fig:interaction_sim}). The one exception is that MVT1 has superior regret to MVT2
when $\alpha_2 = 0$ (Fig.~\ref{fig:interaction_sim}).  This suggests that modeling pairwise interactions is harmful when these interactions 
are not actually present. 

\begin{figure}[h]
\centering
   \includegraphics[width=\columnwidth]{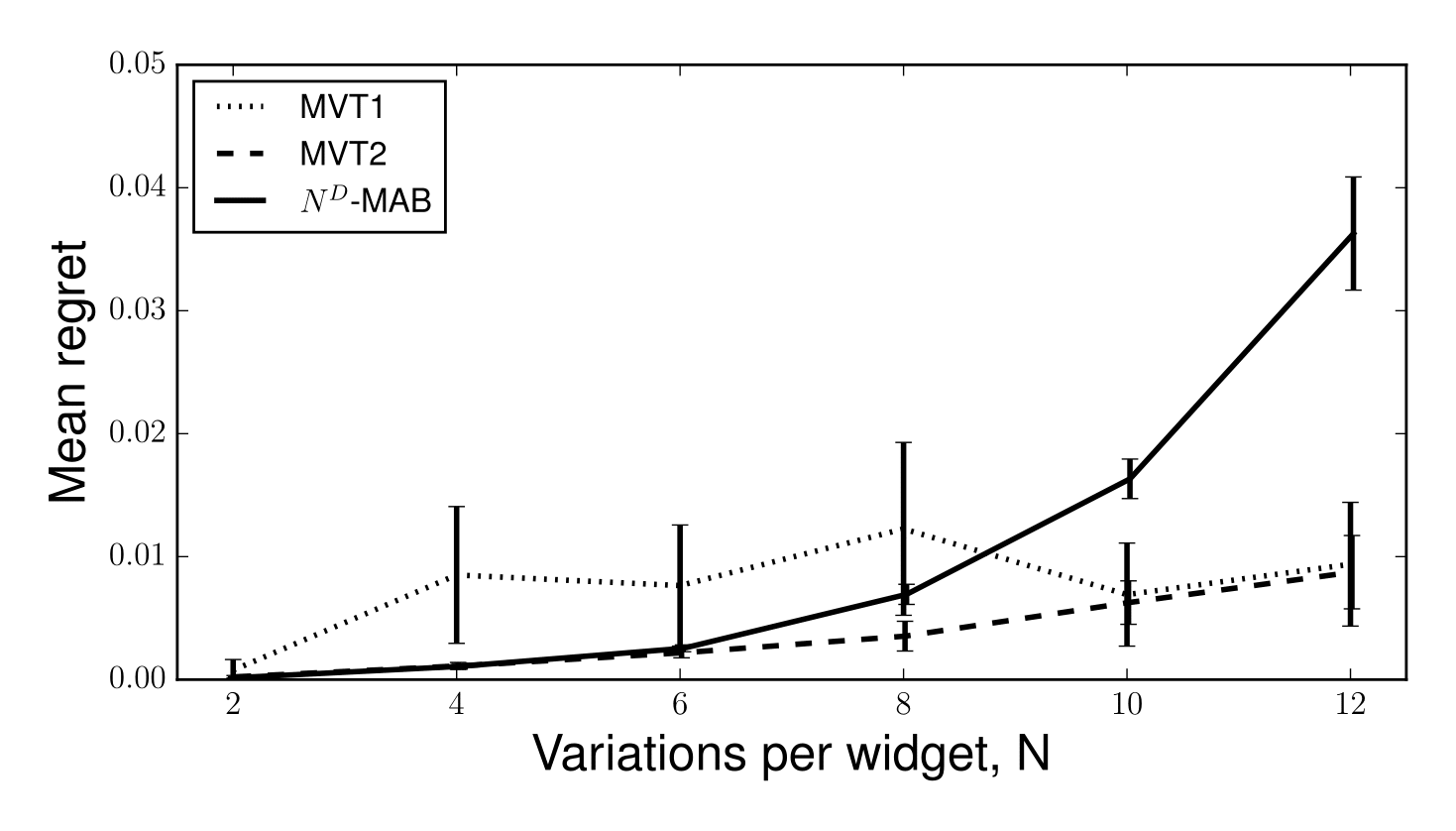}
\caption{Algorithm performance as N is varied.}
\label{fig:complexity_sim}
\end{figure}

Next, we examined the impact of complexity on performance (Fig.~\ref{fig:complexity_sim}). The number of variations per slot, $N$, was systematically varied from 2 to 12. Thus, 
the total number of possible layouts $N^D$ was varied from 8 to 1,728.  For these experiments, we set $\alpha_1=\alpha_2=1$ and $\alpha_c=0$. The $regret$ of all algorithms worsened as the number of layouts grew exponentially. 
However, the $N^D$-MAB algorithm showed the steepest drop in performance with model complexity.  This is because the number of parameters learned by $N^D$-MAB grows exponentially with $N$.  
The shallow decline in the performance of both 
MVT1 and MVT2 as N is increased suggests that either algorithm is appropriate for scenarios involving 1,000s of possible layouts.

\begin{figure}[h]
  \centering
  \includegraphics[width=\columnwidth]{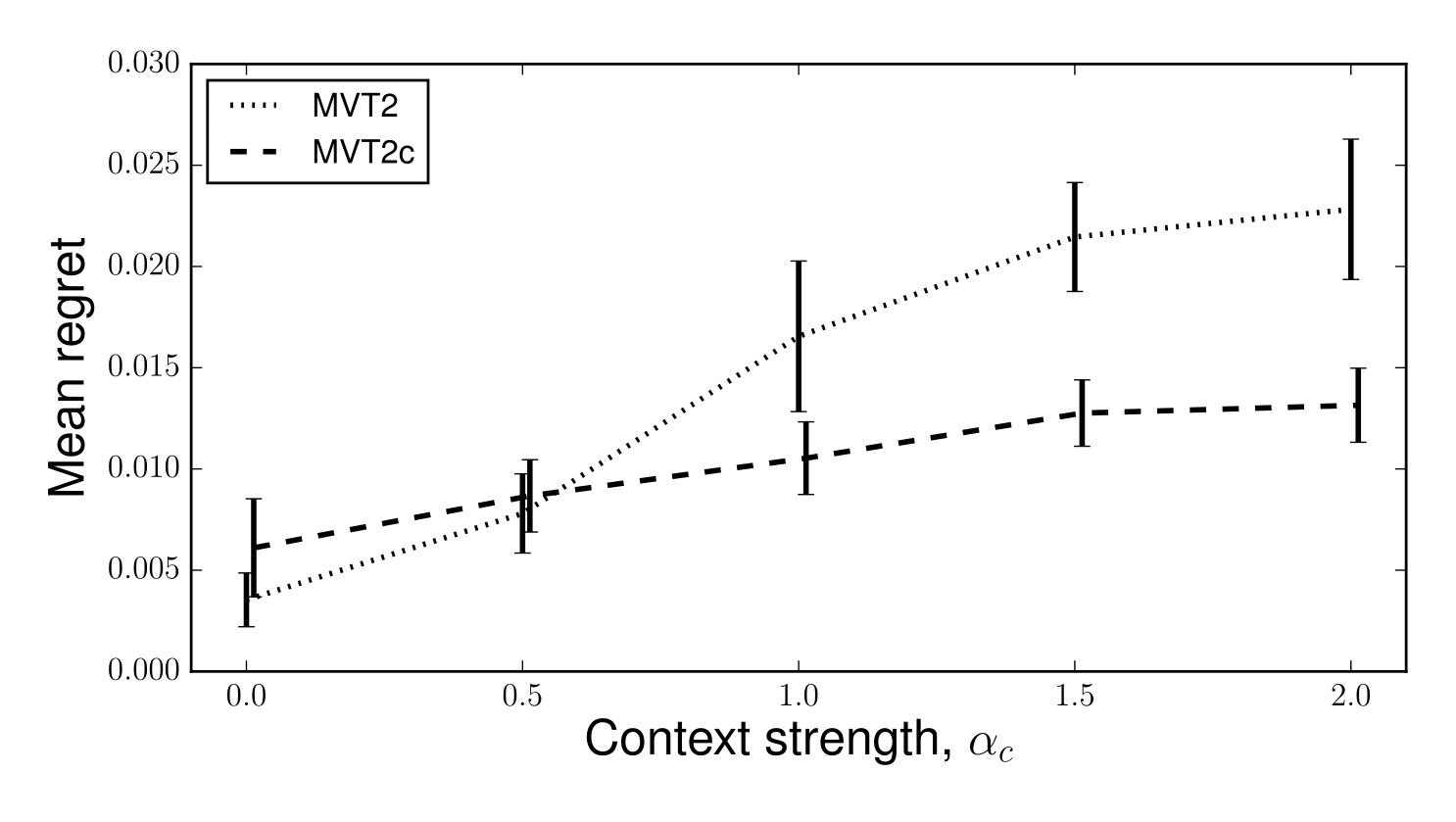}
  \caption{Algorithm performance as $\alpha_c$ is varied.}
  \label{fig:context_sim}
\end{figure}

Finally, we tested the performance of MVT2c in the presence of contextual information and interactions between content and context (Fig.~\ref{fig:context_sim}). We vary the amplitude of $\alpha_c$ while fixing $\alpha_1=\alpha_2=1$. 
To control overall model complexity, we reduce the number of variations so that $N=4$, and use $G=4$ for the number of possible contexts.  
We see superior performance for MVT2c over MVT2 for values of $\alpha_c > 0.5$. This shows that MVT2c lowers regret by accounting for the impact of context; however, when the impact 
of context is very weak, the extra complexity of modeling context may impair performance. 

Taken together, these simulation results suggest that our multivariate bandit algorithm MVT2 is appropriate in scenarios with a large decision space and where interaction effects are present between widgets in the layout. Additionally, the contextual version of our algorithm, MVT2c, performs well in scenarios where the influence of context is significant. Simpler models may show superior regret when these concerns are not in place.

\subsection{Hill climbing impact}
\begin{figure}[h]
\centering
   \begin{subfigure}[b]{\columnwidth}
   \includegraphics[width=\linewidth]{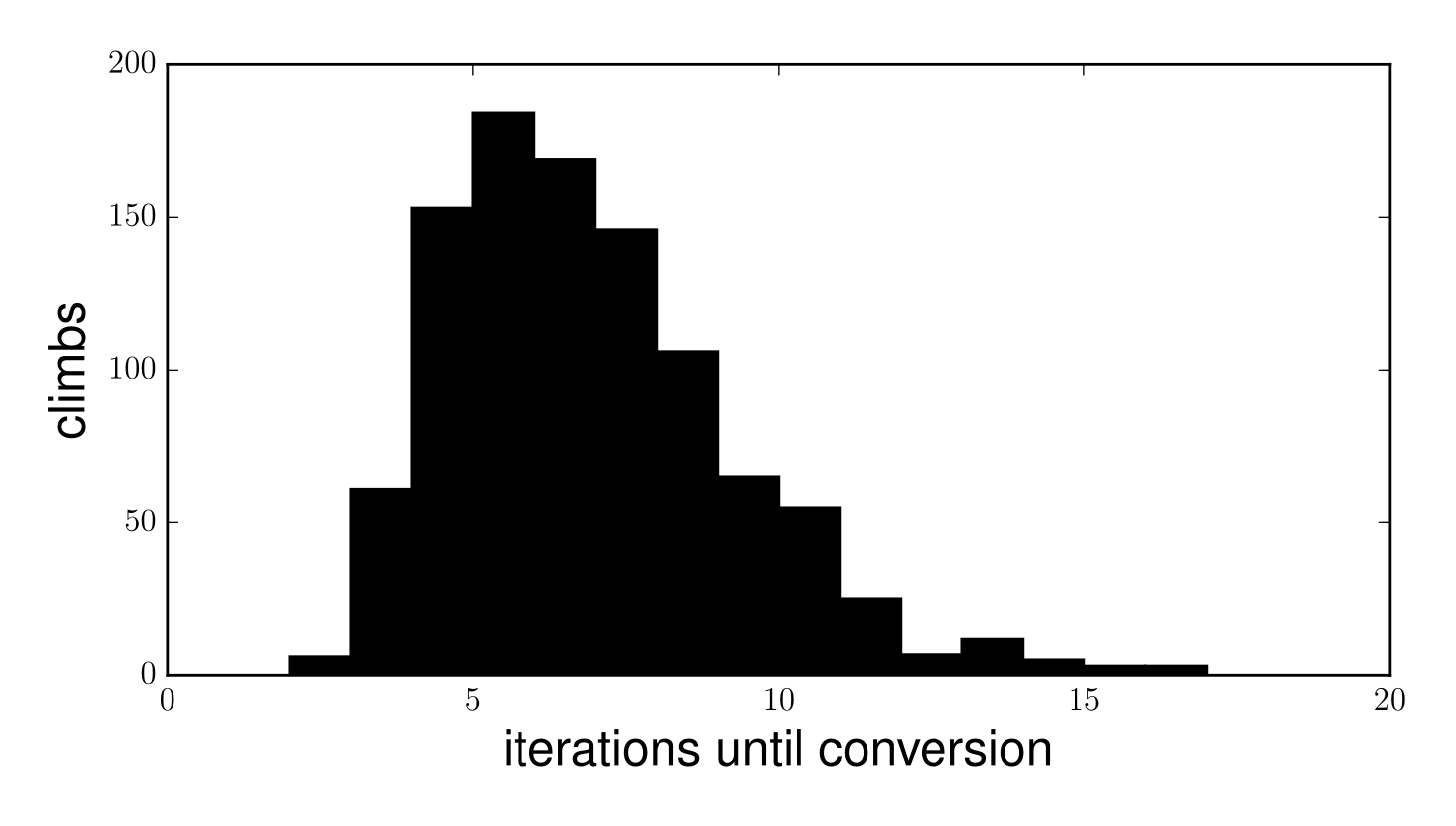}
   \label{fig:hc-sim-histogram} 
\end{subfigure}

   \begin{subfigure}[b]{\columnwidth}
   \includegraphics[width=\linewidth]{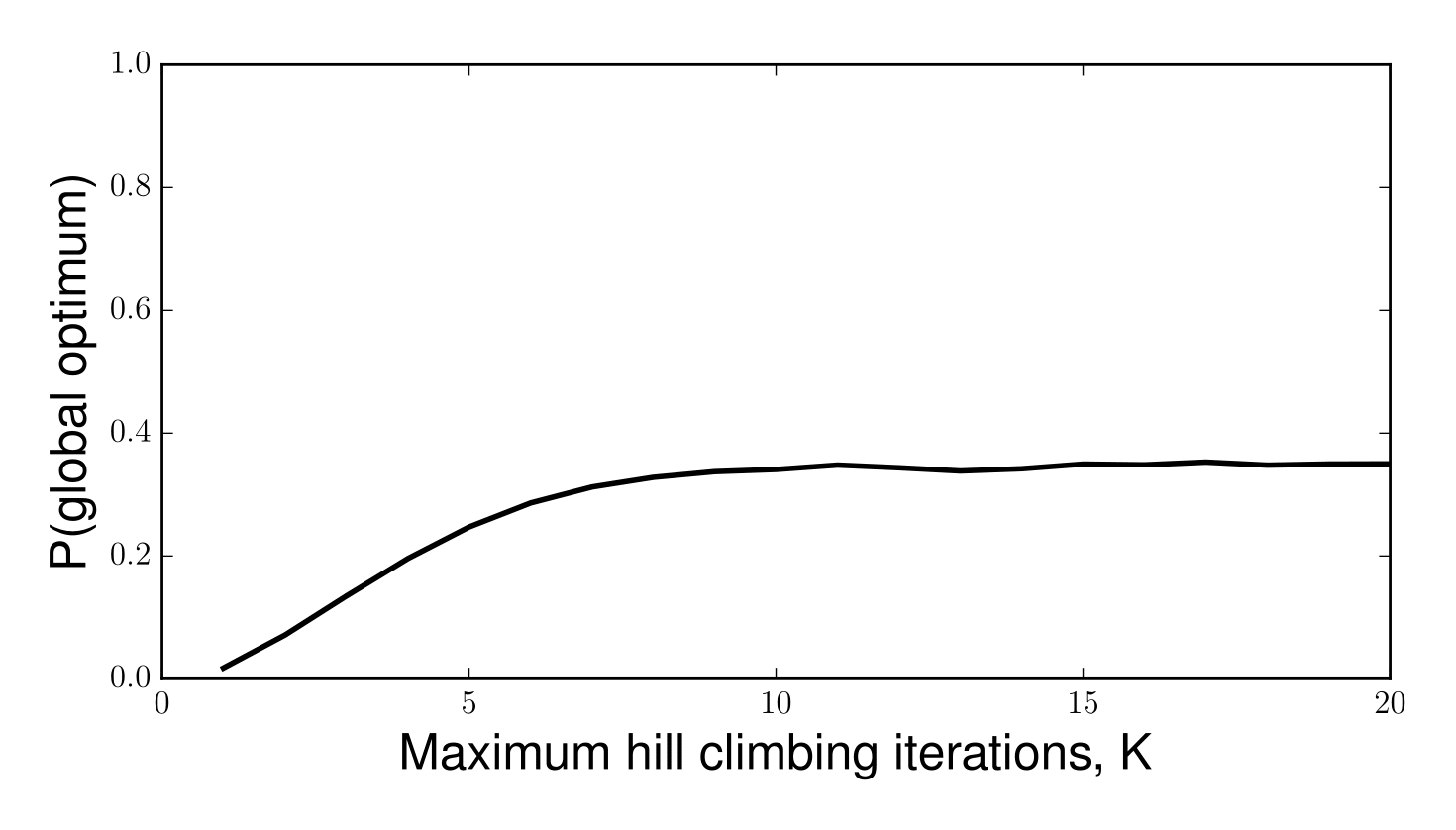}
   \label{fig:hc-sim-pglobal}    
\end{subfigure}

\begin{subfigure}[b]{\columnwidth}
   \includegraphics[width=\linewidth]{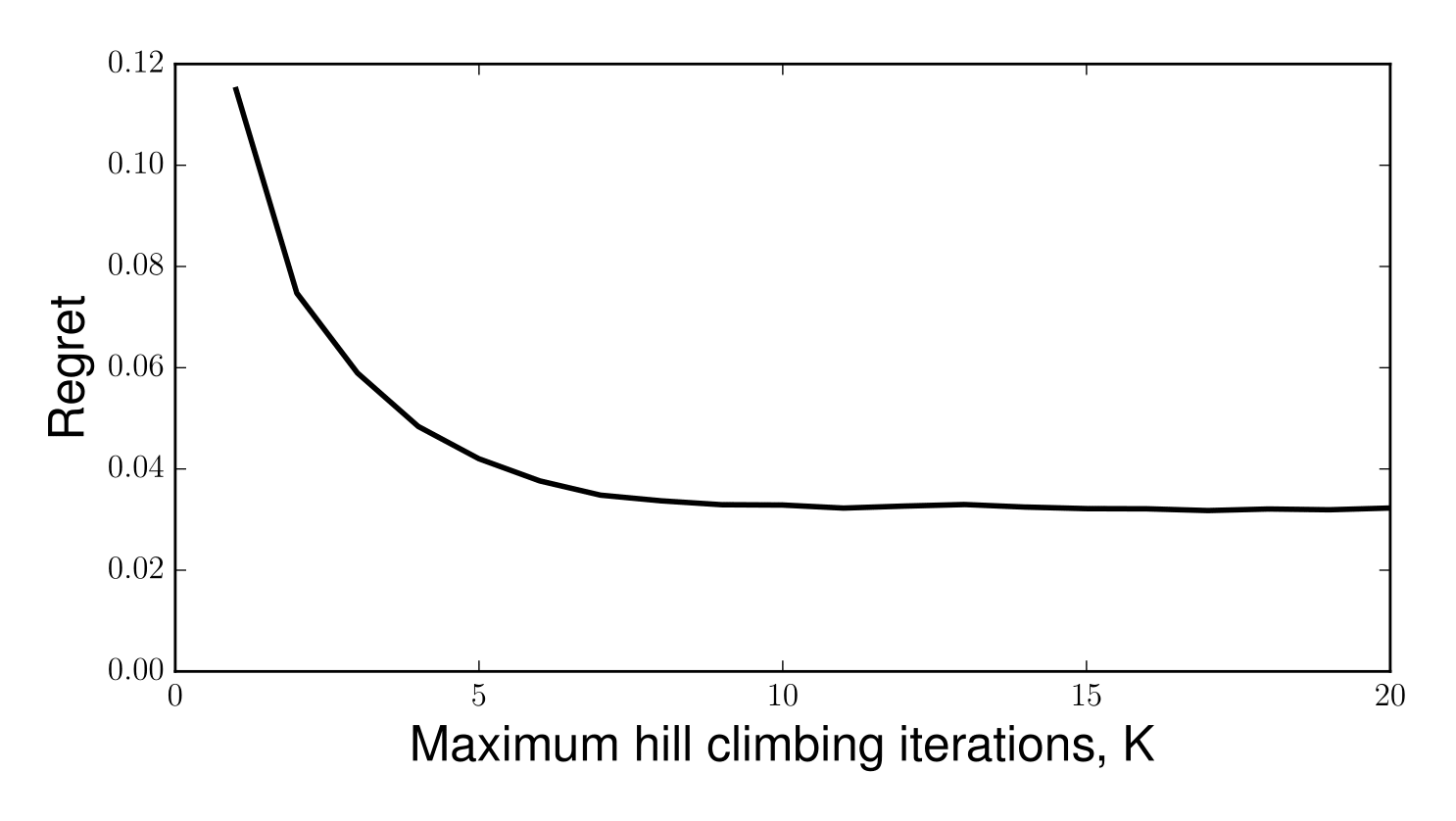}
   \label{fig:hc-sim-regret} 
\end{subfigure}

\caption{Performance of hill climbing as the number of iterations, K, is varied.}
\label{fig:hc-sim}
\end{figure}

We examine the impact of hill climbing on convergence and regret for MVT2. We focus on the scenario where $N=8$, $D=3$, $\alpha_1=\alpha_2=1$, and $\alpha_c=0$. We ran hill climbing algorithm~\ref{alg:HillClimbing} for 1000 times on MVT2 models that were fully trained on instantiations of the simulated data model of equation~\ref{simulator_lm} (see Fig.~\ref{fig:hc-sim}). 

Hill climbing converges quickly for this data set. The number of iterations for hill climbing to converge was $6\pm2.4$ (mean $\pm$ S.D.). This corresponds to only $6 (N-1) + 1 = 42$ distinct layouts being evaluated out of a total of 512 possible layouts. However, this came at the cost that the hill climb reached the global optimum with a probability of $p(global\,optimum) = 0.35$.  
 Despite not always converging to a global optimum, the mean regret was reduced from $0.112$ for a random layout to $0.033$ for the converged solution. We also examined the performance
 of hill climbing as we vary the maximum number of iterations, $K$. The regret and probability of climbing to the global optimum both plateaued at $K=10$. 

Regret can further be reduced by random re-starts. As each run of hill climbing is independent, the probability of reaching the global peak after $S$ restarts is $1 - (1- p(global\,optimum))^S$.  Therefore, with $S=5$, the global optimum can be found with probability greater than $90\%$ after a maximum of $42 * S = 208$ distinct layout evaluations.

\section{Experimental Results}

We now examine the performance of our algorithm on a live production system. 

\subsection{Experimental design}
We applied MVT2 to optimize the layout of a message
promoting the purchase of an Amazon service. The message was shown to a random subset of customers during a visit to Amazon.com on desktop browsers. 
The message was constructed from 5 widgets: a title, an image, a list of bullet points, a thank-you button, and 
a no-thank-you hyperlink (Fig.~\ref{fig:desktop_example}). Each widget could be assigned one of two alternative contents except for the 
image which had three alternatives. The message thus had 48 total variations. 

In addition to MVT2, we included two baseline algorithms in this experiment. The first is $N^D$-MAB which is a multi-armed bandit model with
48 arms (see Eq.~\ref{eq:ts_lm}). The second baseline is a model where each of the $D$ widgets is optimized independently, referred to as D-MABs. Here, each slot $i$ 
is modeled as a separate Bayesian linear probit regression where widget $i$ has linear model:
\begin{equation}
\label{eq:d_mabs}
B_A^TW = W^0 + W^1_i(A).
\end{equation}
The content of widget $i$ is chosen by Thompson sampling on its corresponding model.

Note that the D-MABs algorithm is subtly different from MVT1. While both models ignore interactions between widgets, 
the D-MABs model does not train on a shared error signal. The purpose of including D-MABs in this test was to compare MVT2 performance to a simplistic 
model that does not consider the problem of layout in a combinatorial way. 

During the 12 day experiment, traffic was randomly and equally
distributed between these three algorithms. Traffic consisted of tens of thousands of impressions per algorithm per day. Layouts were selected in real-time using Thompson sampling, and the model was updated once a day after midnight using the previous day's data. Note that only prediction happens online in real-time, while updates are in batch, offline.

\subsection{Analysis Of Results}
\begin{figure}[h]
\centering
\begin{subfigure}[b]{\linewidth}
   \includegraphics[width=\linewidth]{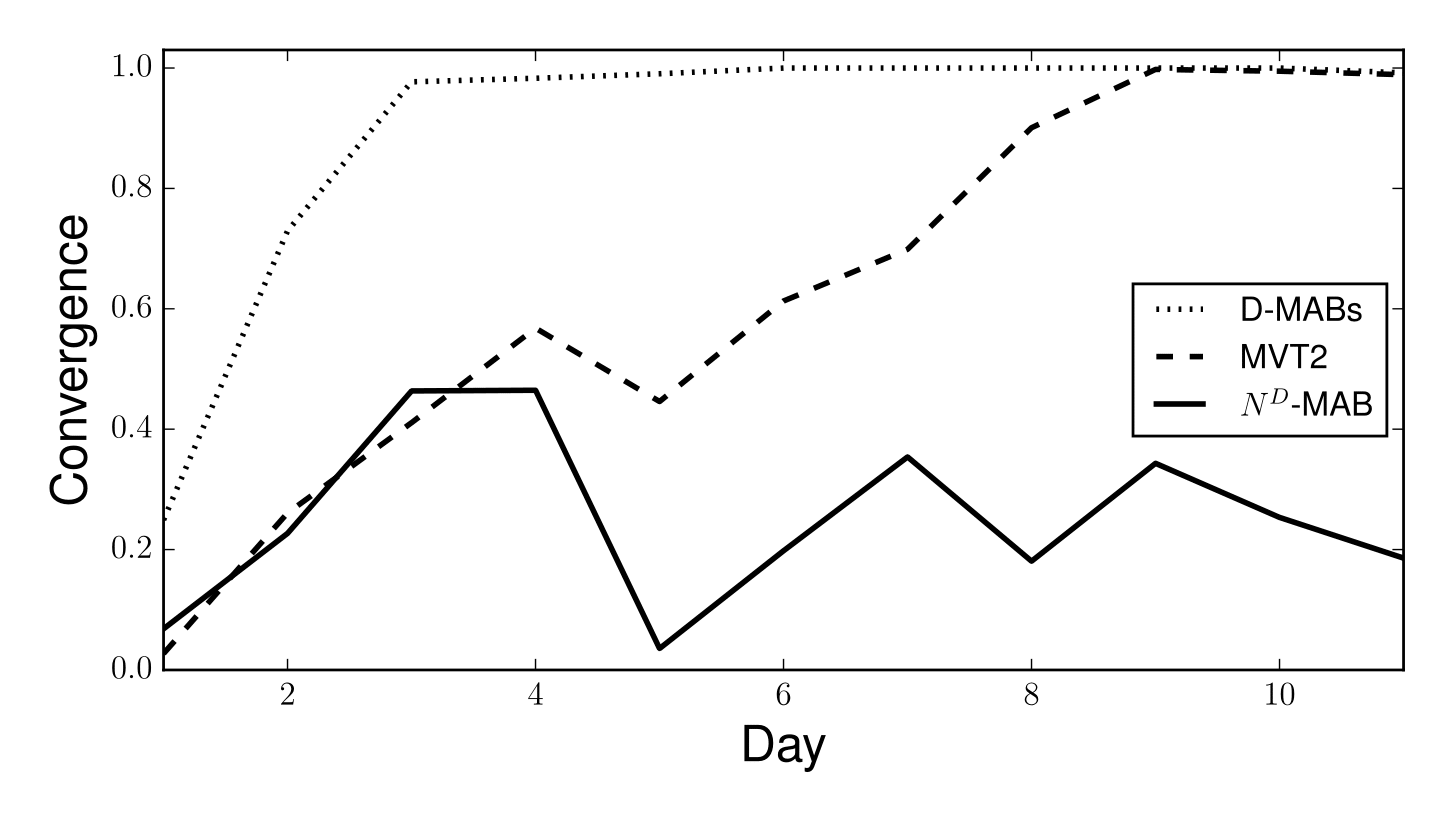}
   \label{fig:sim-ex-conv}
\end{subfigure}
   
\begin{subfigure}[b]{\linewidth}
   \includegraphics[width=\linewidth]{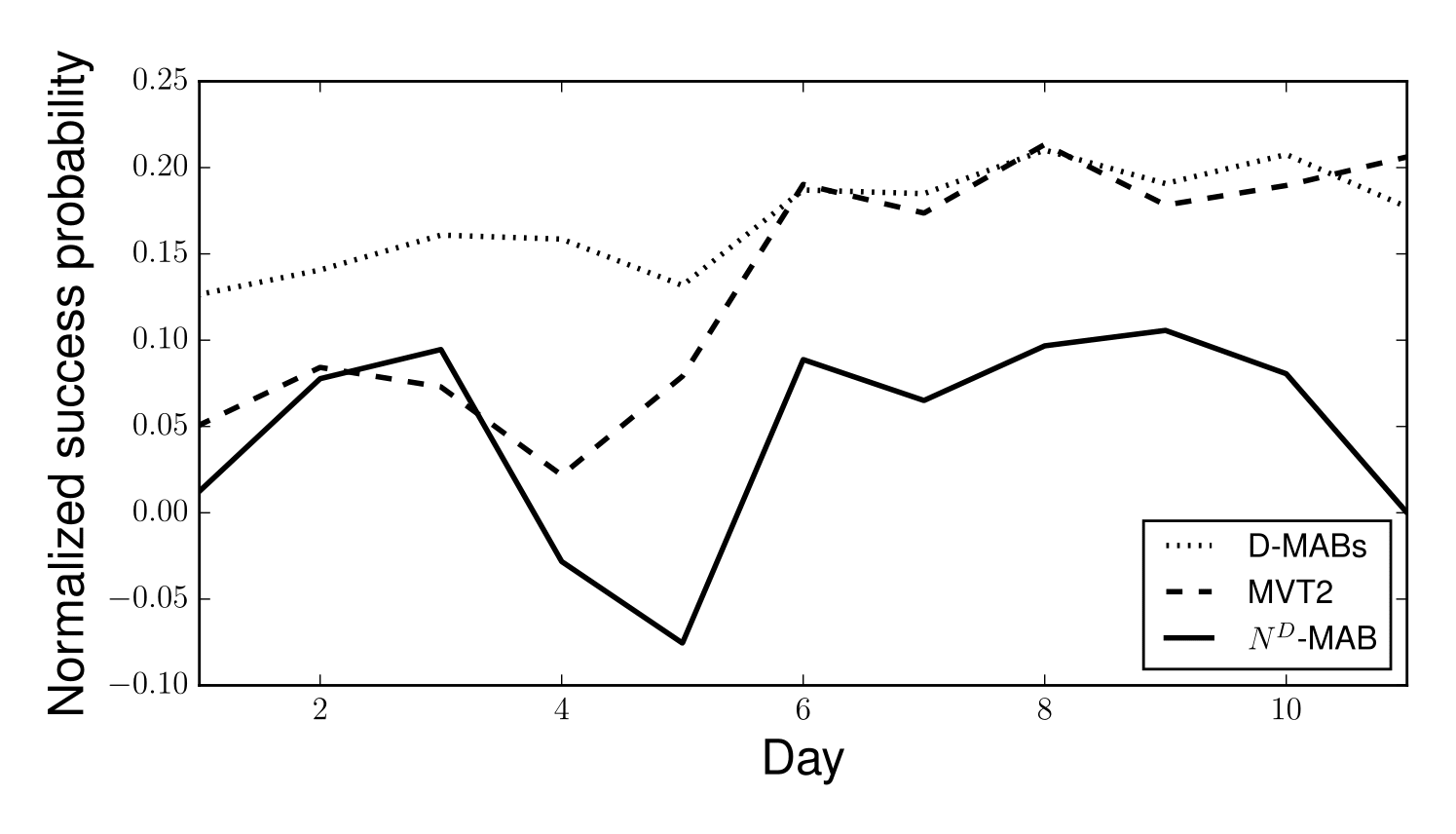}
   \label{fig:sim-ex-regret}
\end{subfigure}

 \begin{subfigure}[b]{\linewidth}
   \includegraphics[width=\linewidth]{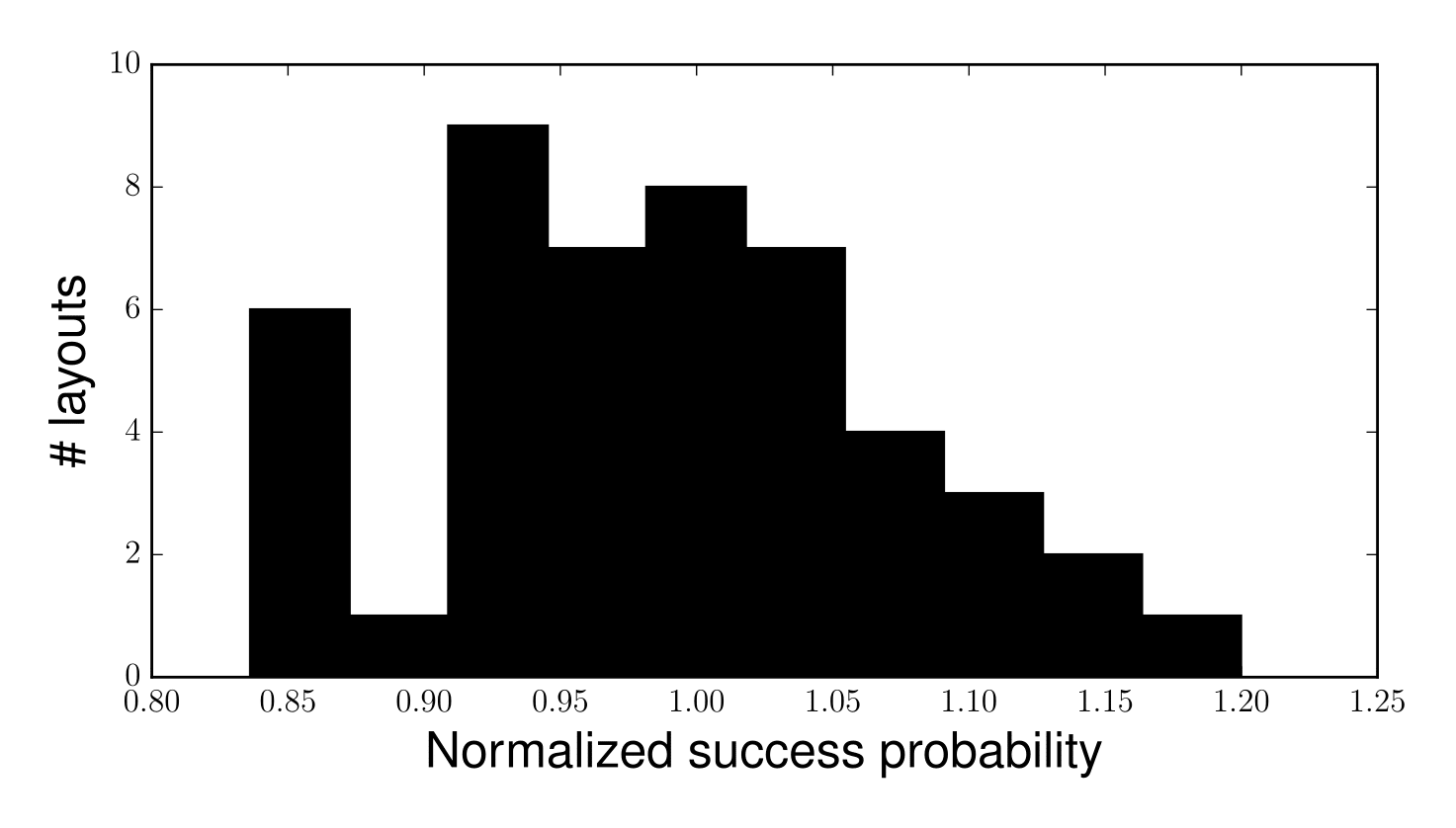}
   \label{fig:sim-ex-histogram} 
\end{subfigure}`

\caption{Results of desktop experiment. Top and middle panels show convergence and normalized performance for each algorithm. Bottom panel shows reward of all layouts normalized by median layout. }
\label{fig:experimental_results}
\end{figure}

Results of the experiment are shown in figure~\ref{fig:experimental_results}. We define the convergence as the proportion of trials on which the algorithm
played its favored layout. A value of 1 means the algorithm always played the same arm and so the model is fully converged. We see that D-MABs converges to a solution in just 3 days followed 
by MVT2 at 9 days. The $N^D$-MAB algorithm shows very little convergence throughout the course of the experiment. While convergence is important for
choosing the optimal layout, a longer convergence period may be tolerated if the regret is low. We plot a normalized success probability as a function of experimental day. For example, 
a normalized success probability of $0.2$ indicates a $20\%$ increase over the performance of the median layout. We see that MVT2's 
success rate is comparable to D-MABs by day 6.

The results from this experiment indicate a non-combinatorial approach can be successful. However, the multivariate MVT2 approach allows us to maintain
robustness for possible interaction effects at only a modest cost for regret and convergence time. The lack of convergence for $N^D$-MAB makes this multi-armed
bandit algorithm inappropriate for fast experimentation.

For comparison with traditional A/B tests, we also performed a 
power analysis for a 48 treatment randomized experiment conditioned on our success rate and traffic size. In order to detect a $5\%$ effect with $p<0.05$ and $\beta<0.20$, 
we estimate that such an experiment would require 66 days~\cite{chow2007sample}.

Finally we note that the winning layout for this experiment showed a $21\%$ lift over the median layout and a $44\%$ lift over the worst
performing layout. If different content had a negligible impact on customer behavior, no algorithmic approach would provide much benefit in optimization. However, given these surprisingly large lifts, there appears to be a large business opportunity in combinatorial optimization of web page layouts.

\subsection{Hill climbing analysis}
\begin{figure}[h]
\centering
   \begin{subfigure}[b]{\linewidth}
   \includegraphics[width=\linewidth]{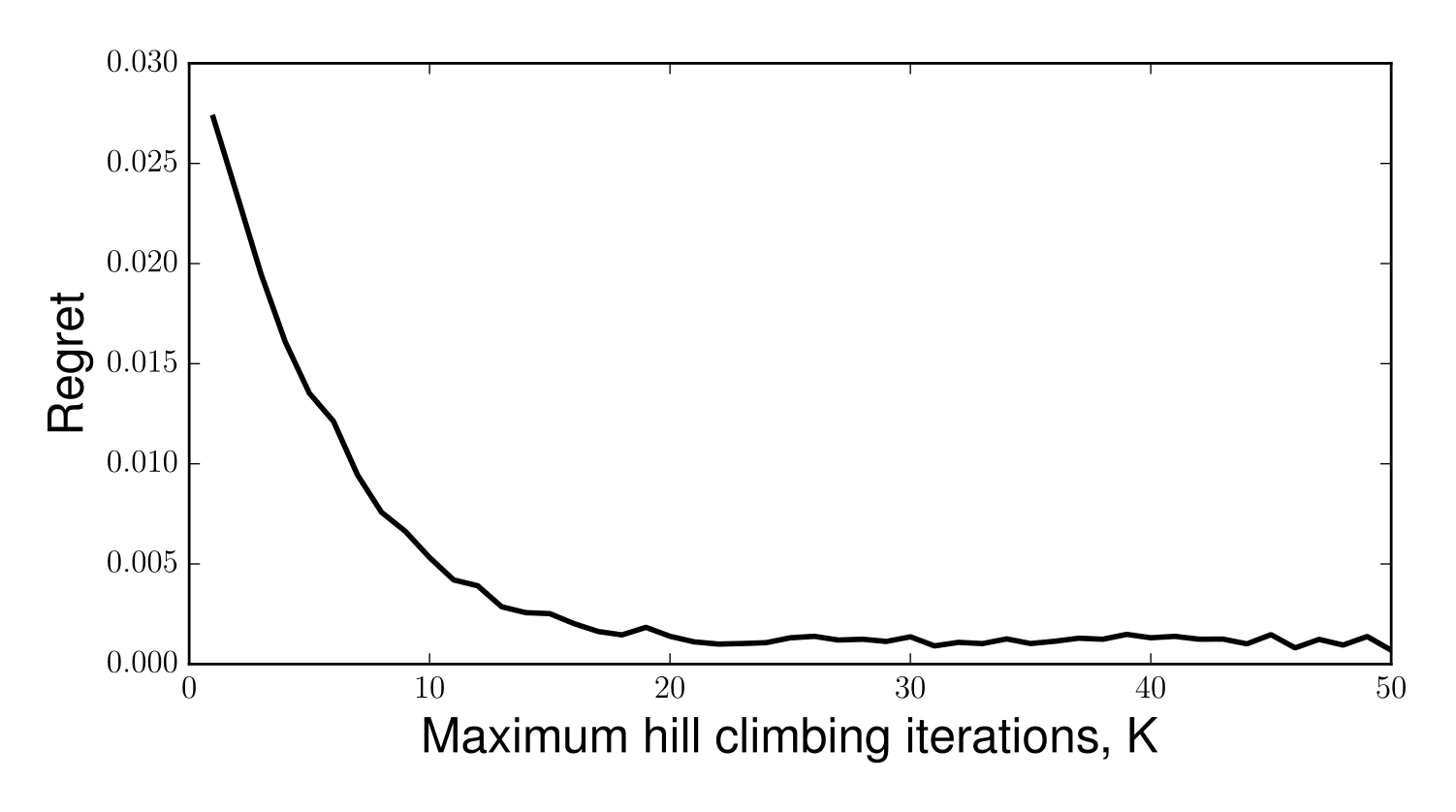}
   \label{fig:sim-ex-regret} 
\end{subfigure}
\begin{subfigure}[b]{\linewidth}
   \includegraphics[width=\linewidth]{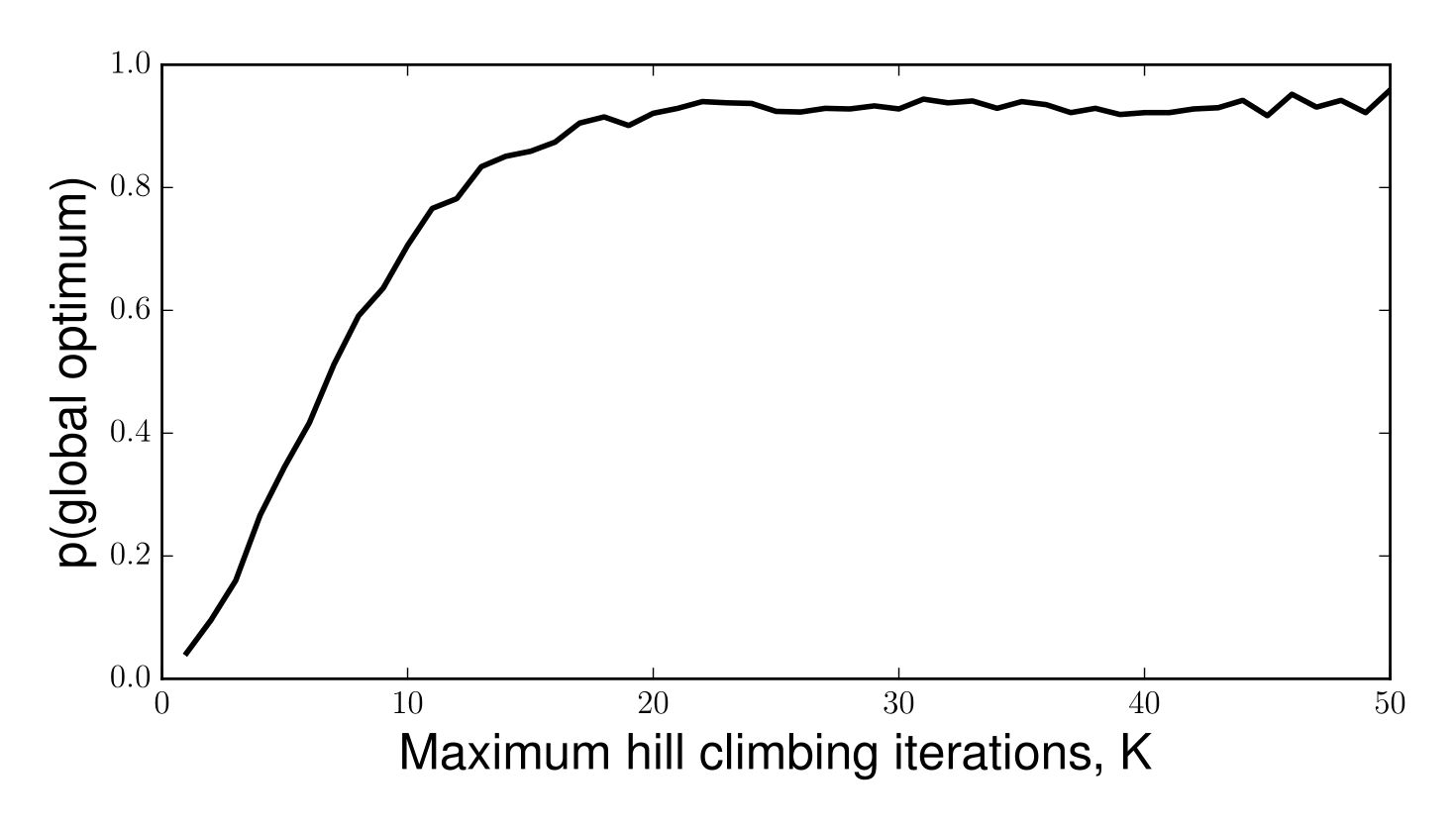}
   \label{fig:sim-ex-conv}
\end{subfigure}
\caption{Regret and probability of identifying global optimum in fully trained model as a function of max iterations for hill-climbing. }
\label{fig:hillclimbing}
\end{figure}

The MVT2 algorithm in this experiment used hill climbing with $S=5$ random restarts and a maximum of $K=48$ iterations to choose which layout to display. To better understand the search performance on a model trained on real data,
we ran simulations of hill-climbing for this fully-trained on real-data MVT2 model (Fig.~\ref{fig:hillclimbing}). With 1000 simulations and no restarts, hill-climbing converged to a local optimum after 24 iterations on average, with $p(global\,optimum)=0.937$.
This probability is greater than in the simulation experiments, suggesting real-world data may have fewer local optima. If we had set $K=15$ and $S=2$ , then the algorithm would have achieved $p(global\,optimum)=0.95$ at $50\%$ of the effort of an exhaustive search.

\subsection{Widget interactions}
The fact that both MVT2 and D-MABs converged to the same solution suggests that interactions between content may have been weak in our experiment. 
We verified this through a likelihood ratio test~\cite{casella2002statistical} comparing goodness-of-fit for models with different levels of interaction between widgets. In addition to MVT2, we
tested two variations with different interaction levels: MVT1 with no pairwise interaction terms, and MVT3 with additional weights for each of the ${D \choose 3}N^3$ possible third-order (three-way) content interaction. Each model was trained on production data where the templates were shown to customers uniformly at random. The likelihood ratio test 
applied to MVT2 versus MVT1 and MVT3 versus MVT2 were both insignificant ($p > 0.05$). 

While we did not find significant 2nd or 3rd order interactions in this
experiment, this observation does not generalize. In a follow-up experiment, we applied MVT2 to the mobile version of the promotional page.
The template for the mobile page consists of 5 widgets with 2 alternatives each for a total of 32 possible layouts. In this case, we found that 2nd order ($p < 0.01$) but not 3rd order
effects were significant (Table~\ref{table:interactions}). As it is often hard to determine in advance the degree of interaction for a particular experiment, one may want to at least include 2nd order interactions, and opt for a multivariate bandit formulation. 

\begin{table}[h]
    \caption{p-value of interaction effects in production experiments. * denotes significance.}
    \label{table:interactions}
  \begin{center}
    \begin{tabular}{|| l | c | c ||} \hline
      \textbf{Interaction} &  \textbf{Desktop} & \textbf{Mobile} \\ \hline
     \textbf{2nd-order}  & 0.2577 & 0.0096* \\ \hline
     \textbf{3rd-order} & 0.428 & 0.1735 \\ \hline
    \end{tabular}
  \end{center}
\end{table}

\section{Discussion And Conclusions}
We present an algorithm for performing multivariate optimization on large
decision spaces. Simulation results show that our algorithm scales well to problems 
involving 1,000s of layouts, converging in a practical amount of time.  We have shown 
the suitability of our method for capturing interactions between content that are displayed 
together while also accounting for the effects of context. 
Our algorithm balances exploration with exploitation during learning, allowing for continuous optimization. It also searches efficiently 
through an exponential layout space in realtime.
We have applied this approach to promoting purchases of an Amazon service where we saw a large business impact after only a single week of experimentation.
We are actively deploying this algorithm to other areas of Amazon.com to solve increasingly complex decision problems.

Several extensions to this work could potentially enhance its performance. One limitation of the current framework 
is that the widget contents are represented as identifiers. This prevents generalization between
related creatives, such as small variations on a common theme or message. It is straight-forward to featurize
content so that the model can learn what properties of the content are most important. This could include 
adding positional features relevant to how users browse two-dimensional web pages~\cite{chierichetti2011optimizing}. Second, we note that higher order 
input features such as widget interactions should be regularized more heavily than lower order features to
reduce parametric complexity. The complexity of representing interactions could also be reduced through techniques such as factorization machines~\cite{rendle2010factorization}
and neural networks~\cite{bengio2013representation} that produce low-dimension embeddings out of high-order features. Finally, it still remains to understand
the impact of using hill climbing to approximate Thompson sampling. How does this procedure impact regret? One could establish a regret bound that is then
tightened through search strategy refinements.

More generally, our algorithm can be applied to any problem that involves combinatorial decision making.
It allows for experimental throughput and continuous learning that is out of reach for traditional randomized 
experiments. Furthermore, by using a parametric modeling approach, we allow business insights to be extracted directly from the 
parameters of the trained model. Alternative content for a web page is typically inexpensive to produce,
and this algorithm allows for fast filtering-out of poor choices. Multivariate optimization may thus encourage the
exploration of riskier and more creative approaches in the creation of online content.

\begin{acks}
The authors thank Charles Elkan, Sriram Srinavasan, Milos Curcic, Andrea Qualizza, Sham Kakade, Karthik Mohan, and Tao Hu for their helpful discussions.
\end{acks}

\bibliographystyle{ACM-Reference-Format}
\balance 
\bibliography{s../mvt} 


\begin{thebibliography}{00}


\ifx \showCODEN    \undefined \def \showCODEN     #1{\unskip}     \fi
\ifx \showDOI      \undefined \def \showDOI       #1{{\tt DOI:}\penalty0{#1}\ }
  \fi
\ifx \showISBNx    \undefined \def \showISBNx     #1{\unskip}     \fi
\ifx \showISBNxiii \undefined \def \showISBNxiii  #1{\unskip}     \fi
\ifx \showISSN     \undefined \def \showISSN      #1{\unskip}     \fi
\ifx \showLCCN     \undefined \def \showLCCN      #1{\unskip}     \fi
\ifx \shownote     \undefined \def \shownote      #1{#1}          \fi
\ifx \showarticletitle \undefined \def \showarticletitle #1{#1}   \fi
\ifx \showURL      \undefined \def \showURL       #1{#1}          \fi
\providecommand\bibfield[2]{#2}
\providecommand\bibinfo[2]{#2}
\providecommand\natexlab[1]{#1}
\providecommand\showeprint[2][]{arXiv:#2}

\bibitem[\protect\citeauthoryear{Agrawal and Goyal}{Agrawal and Goyal}{2012}]%
        {agrawal2012analysis}
\bibfield{author}{\bibinfo{person}{Shipra Agrawal} {and} \bibinfo{person}{Navin
  Goyal}.} \bibinfo{year}{2012}\natexlab{}.
\newblock \showarticletitle{Analysis of Thompson Sampling for the Multi-armed
  Bandit Problem.}. In \bibinfo{booktitle}{{\em COLT}}. \bibinfo{pages}{39--1}.
\newblock


\bibitem[\protect\citeauthoryear{Agrawal and Goyal}{Agrawal and Goyal}{2013}]%
        {Agrawal2013ContextualTSBounds}
\bibfield{author}{\bibinfo{person}{Shipra Agrawal} {and} \bibinfo{person}{Navin
  Goyal}.} \bibinfo{year}{2013}\natexlab{}.
\newblock \showarticletitle{Thompson Sampling for Contextual Bandits with
  Linear Payoffs}. In \bibinfo{booktitle}{{\em Proceedings of the 30th
  International Conference on Machine Learning (ICML)}}.
  \bibinfo{publisher}{JMLR}, \bibinfo{address}{Atlanta, Georgia},
  \bibinfo{pages}{127--135}.
\newblock


\bibitem[\protect\citeauthoryear{Bengio, Courville, and Vincent}{Bengio
  et~al\mbox{.}}{2013}]%
        {bengio2013representation}
\bibfield{author}{\bibinfo{person}{Yoshua Bengio}, \bibinfo{person}{Aaron
  Courville}, {and} \bibinfo{person}{Pascal Vincent}.}
  \bibinfo{year}{2013}\natexlab{}.
\newblock \showarticletitle{Representation learning: A review and new
  perspectives}.
\newblock \bibinfo{journal}{{\em IEEE transactions on pattern analysis and
  machine intelligence\/}} \bibinfo{volume}{35}, \bibinfo{number}{8}
  (\bibinfo{year}{2013}), \bibinfo{pages}{1798--1828}.
\newblock


\bibitem[\protect\citeauthoryear{Box, Hunter, and Hunter}{Box
  et~al\mbox{.}}{2005}]%
        {box2005statistics}
\bibfield{author}{\bibinfo{person}{George~EP Box}, \bibinfo{person}{J~Stuart
  Hunter}, {and} \bibinfo{person}{William~Gordon Hunter}.}
  \bibinfo{year}{2005}\natexlab{}.
\newblock \bibinfo{booktitle}{{\em Statistics for experimenters: design,
  innovation, and discovery}}. Vol.~\bibinfo{volume}{2}.
\newblock \bibinfo{publisher}{Wiley-Interscience New York}.
\newblock


\bibitem[\protect\citeauthoryear{Bubeck and Cesa-Bianchi}{Bubeck and
  Cesa-Bianchi}{2012}]%
        {bubeck2012BanditsSurvey}
\bibfield{author}{\bibinfo{person}{Sebastien Bubeck} {and}
  \bibinfo{person}{Nicolo Cesa-Bianchi}.} \bibinfo{year}{2012}\natexlab{}.
\newblock \showarticletitle{Regret Analysis of Stochastic and Nonstochastic
  Multi-armed Bandit Problems}.
\newblock \bibinfo{journal}{{\em Foundations and Trends in Machine learning\/}}
  \bibinfo{volume}{5}, \bibinfo{number}{1} (\bibinfo{year}{2012}),
  \bibinfo{pages}{1--122}.
\newblock


\bibitem[\protect\citeauthoryear{Casella and Berger}{Casella and
  Berger}{2002}]%
        {casella2002statistical}
\bibfield{author}{\bibinfo{person}{George Casella} {and}
  \bibinfo{person}{Roger~L Berger}.} \bibinfo{year}{2002}\natexlab{}.
\newblock \bibinfo{booktitle}{{\em Statistical inference}}.
  Vol.~\bibinfo{volume}{2}.
\newblock \bibinfo{publisher}{Duxbury Pacific Grove, CA}.
\newblock


\bibitem[\protect\citeauthoryear{Cesa-Bianchi and Lugosi}{Cesa-Bianchi and
  Lugosi}{2012}]%
        {cesa2012combinatorial}
\bibfield{author}{\bibinfo{person}{Nicol{\`o} Cesa-Bianchi} {and}
  \bibinfo{person}{G{\'a}bor Lugosi}.} \bibinfo{year}{2012}\natexlab{}.
\newblock \showarticletitle{Combinatorial bandits}.
\newblock \bibinfo{journal}{{\it J. Comput. System Sci.}} \bibinfo{volume}{78},
  \bibinfo{number}{5} (\bibinfo{year}{2012}), \bibinfo{pages}{1404--1422}.
\newblock


\bibitem[\protect\citeauthoryear{Chapelle and Li}{Chapelle and Li}{2011}]%
        {chapelle2011empirical}
\bibfield{author}{\bibinfo{person}{Olivier Chapelle} {and}
  \bibinfo{person}{Lihong Li}.} \bibinfo{year}{2011}\natexlab{}.
\newblock \showarticletitle{An empirical evaluation of thompson sampling}. In
  \bibinfo{booktitle}{{\em Advances in neural information processing systems}}.
  \bibinfo{pages}{2249--2257}.
\newblock


\bibitem[\protect\citeauthoryear{Chierichetti, Kumar, and
  Raghavan}{Chierichetti et~al\mbox{.}}{2011}]%
        {chierichetti2011optimizing}
\bibfield{author}{\bibinfo{person}{Flavio Chierichetti}, \bibinfo{person}{Ravi
  Kumar}, {and} \bibinfo{person}{Prabhakar Raghavan}.}
  \bibinfo{year}{2011}\natexlab{}.
\newblock \showarticletitle{Optimizing two-dimensional search results
  presentation}. In \bibinfo{booktitle}{{\em Proceedings of the fourth ACM
  international conference on Web search and data mining}}. ACM,
  \bibinfo{pages}{257--266}.
\newblock


\bibitem[\protect\citeauthoryear{Chow, Wang, and Shao}{Chow
  et~al\mbox{.}}{2007}]%
        {chow2007sample}
\bibfield{author}{\bibinfo{person}{Shein-Chung Chow}, \bibinfo{person}{Hansheng
  Wang}, {and} \bibinfo{person}{Jun Shao}.} \bibinfo{year}{2007}\natexlab{}.
\newblock \bibinfo{booktitle}{{\em Sample size calculations in clinical
  research}}.
\newblock \bibinfo{publisher}{CRC press}.
\newblock


\bibitem[\protect\citeauthoryear{Dani, Hayes, and Kakade}{Dani
  et~al\mbox{.}}{2008}]%
        {Dani2008ContextualBanditBound}
\bibfield{author}{\bibinfo{person}{Varsha Dani}, \bibinfo{person}{Thomas~P.
  Hayes}, {and} \bibinfo{person}{Sham~M. Kakade}.}
  \bibinfo{year}{2008}\natexlab{}.
\newblock \showarticletitle{Stochastic Linear Optimization under Bandit
  Feedback}. In \bibinfo{booktitle}{{\em Proceedings of the 21st Annual
  Conference on Learning Theory (COLT)}}. \bibinfo{address}{Helsinki, Finland},
  \bibinfo{pages}{355--366}.
\newblock


\bibitem[\protect\citeauthoryear{Graepel, Candela, Borchert, and
  Herbrich}{Graepel et~al\mbox{.}}{2010}]%
        {graepel2010web}
\bibfield{author}{\bibinfo{person}{Thore Graepel}, \bibinfo{person}{Joaquin~Q
  Candela}, \bibinfo{person}{Thomas Borchert}, {and} \bibinfo{person}{Ralf
  Herbrich}.} \bibinfo{year}{2010}\natexlab{}.
\newblock \showarticletitle{Web-scale bayesian click-through rate prediction
  for sponsored search advertising in microsoft's bing search engine}. In
  \bibinfo{booktitle}{{\em Proceedings of International Conference on Machine
  Learning (ICML)}}. \bibinfo{address}{Haifa, Israel}, \bibinfo{pages}{13--20}.
\newblock


\bibitem[\protect\citeauthoryear{Joseph}{Joseph}{2006}]%
        {joseph2006experiments}
\bibfield{author}{\bibinfo{person}{V~Roshan Joseph}.}
  \bibinfo{year}{2006}\natexlab{}.
\newblock \showarticletitle{Experiments: Planning, Analysis, and Parameter
  Design Optimization}.
\newblock \bibinfo{journal}{{\em IIE Transactions\/}} \bibinfo{volume}{38},
  \bibinfo{number}{6} (\bibinfo{year}{2006}), \bibinfo{pages}{521--523}.
\newblock


\bibitem[\protect\citeauthoryear{Li, Chu, Langford, and Schapire}{Li
  et~al\mbox{.}}{2010}]%
        {li2010contextual}
\bibfield{author}{\bibinfo{person}{Lihong Li}, \bibinfo{person}{Wei Chu},
  \bibinfo{person}{John Langford}, {and} \bibinfo{person}{Robert~E Schapire}.}
  \bibinfo{year}{2010}\natexlab{}.
\newblock \showarticletitle{A contextual-bandit approach to personalized news
  article recommendation}. In \bibinfo{booktitle}{{\em Proceedings of the 19th
  international conference on World wide web}}. ACM, \bibinfo{pages}{661--670}.
\newblock


\bibitem[\protect\citeauthoryear{Macambira and {de Souza}}{Macambira and {de
  Souza}}{2000}]%
        {macambira2000edgeWeightedClique}
\bibfield{author}{\bibinfo{person}{Elder~Magalh{\~a}es Macambira} {and}
  \bibinfo{person}{Cid~Carvalho {de Souza}}.} \bibinfo{year}{2000}\natexlab{}.
\newblock \showarticletitle{The edge-weighted clique problem: Valid
  inequalities, facets and polyhedral computations}.
\newblock \bibinfo{journal}{{\em European Journal of Operational Research\/}}
  \bibinfo{volume}{123}, \bibinfo{number}{2} (\bibinfo{year}{2000}),
  \bibinfo{pages}{346--371}.
\newblock


\bibitem[\protect\citeauthoryear{Mohan and Dekel}{Mohan and Dekel}{2011}]%
        {mohan2011bipartite}
\bibfield{author}{\bibinfo{person}{Karthik Mohan} {and} \bibinfo{person}{Ofer
  Dekel}.} \bibinfo{year}{2011}\natexlab{}.
\newblock \showarticletitle{Online bipartite matching with partially-bandit
  feedback}. In \bibinfo{booktitle}{{\em Proceedings of NIPS workshop on
  Discrete optimization in Machine Learning}}. \bibinfo{address}{Granada,
  Spain}, \bibinfo{pages}{1--7}.
\newblock


\bibitem[\protect\citeauthoryear{Nassif, Cansizlar, Goodman, and
  Vishwanathan}{Nassif et~al\mbox{.}}{2016}]%
        {nassif2016music}
\bibfield{author}{\bibinfo{person}{Houssam Nassif}, \bibinfo{person}{Kemal~Oral
  Cansizlar}, \bibinfo{person}{Mitchell Goodman}, {and}
  \bibinfo{person}{S.~V.~N. Vishwanathan}.} \bibinfo{year}{2016}\natexlab{}.
\newblock \showarticletitle{Diversifying Music Recommendations}. In
  \bibinfo{booktitle}{{\em Proceedings of Machine Learning for Music Discovery
  Workshop at $33^{rd}$ International Conference on Machine Learning (ICML)}}.
\newblock


\bibitem[\protect\citeauthoryear{Rendle}{Rendle}{2010}]%
        {rendle2010factorization}
\bibfield{author}{\bibinfo{person}{Steffen Rendle}.}
  \bibinfo{year}{2010}\natexlab{}.
\newblock \showarticletitle{Factorization machines}. In
  \bibinfo{booktitle}{{\em Data Mining (ICDM), 2010 IEEE 10th International
  Conference on}}. IEEE, \bibinfo{pages}{995--1000}.
\newblock


\bibitem[\protect\citeauthoryear{Tang, Rosales, Singh, and Agarwal}{Tang
  et~al\mbox{.}}{2013}]%
        {tang2013automatic}
\bibfield{author}{\bibinfo{person}{Liang Tang}, \bibinfo{person}{Romer
  Rosales}, \bibinfo{person}{Ajit Singh}, {and} \bibinfo{person}{Deepak
  Agarwal}.} \bibinfo{year}{2013}\natexlab{}.
\newblock \showarticletitle{Automatic ad format selection via contextual
  bandits}. In \bibinfo{booktitle}{{\em Proceedings of the 22nd ACM
  international conference on Conference on information \& knowledge
  management}}. ACM, \bibinfo{pages}{1587--1594}.
\newblock


\bibitem[\protect\citeauthoryear{Teo, Nassif, Hill, Srinivasan, Goodman, Mohan,
  and Vishwanathan}{Teo et~al\mbox{.}}{2016}]%
        {Teo2016airstream}
\bibfield{author}{\bibinfo{person}{Choon~Hui Teo}, \bibinfo{person}{Houssam
  Nassif}, \bibinfo{person}{Daniel Hill}, \bibinfo{person}{Sriram Srinivasan},
  \bibinfo{person}{Mitchell Goodman}, \bibinfo{person}{Vijai Mohan}, {and}
  \bibinfo{person}{S.V.N. Vishwanathan}.} \bibinfo{year}{2016}\natexlab{}.
\newblock \showarticletitle{Adaptive, Personalized Diversity for Visual
  Discovery}. In \bibinfo{booktitle}{{\em Proceedings of the 10th ACM
  Conference on Recommender Systems (RecSys)}}. \bibinfo{publisher}{ACM},
  \bibinfo{address}{Boston}, \bibinfo{pages}{35--38}.
\newblock


\bibitem[\protect\citeauthoryear{Wang, Yin, Jie, Wang, Yamada, Chang, and
  Mei}{Wang et~al\mbox{.}}{2016}]%
        {wang2016beyond}
\bibfield{author}{\bibinfo{person}{Yue Wang}, \bibinfo{person}{Dawei Yin},
  \bibinfo{person}{Luo Jie}, \bibinfo{person}{Pengyuan Wang},
  \bibinfo{person}{Makoto Yamada}, \bibinfo{person}{Yi Chang}, {and}
  \bibinfo{person}{Qiaozhu Mei}.} \bibinfo{year}{2016}\natexlab{}.
\newblock \showarticletitle{Beyond ranking: Optimizing whole-page
  presentation}. In \bibinfo{booktitle}{{\em Proceedings of the Ninth ACM
  International Conference on Web Search and Data Mining}}. ACM,
  \bibinfo{pages}{103--112}.
\newblock


\bibitem[\protect\citeauthoryear{Yue and Guestrin}{Yue and Guestrin}{2011}]%
        {yue2011linear}
\bibfield{author}{\bibinfo{person}{Yisong Yue} {and} \bibinfo{person}{Carlos
  Guestrin}.} \bibinfo{year}{2011}\natexlab{}.
\newblock \showarticletitle{Linear submodular bandits and their application to
  diversified retrieval}. In \bibinfo{booktitle}{{\em Advances in Neural
  Information Processing Systems}}. \bibinfo{pages}{2483--2491}.
\newblock


\end{thebibliography}

\end{document}